\documentclass{article}

\usepackage[preprint]{neurips_2026}
\usepackage{graphicx}

\usepackage[utf8]{inputenc} 
\usepackage[T1]{fontenc}    
\usepackage{hyperref}       
\usepackage{url}            
\usepackage{booktabs}       
\usepackage{amsfonts}       
\usepackage{nicefrac}       
\usepackage{microtype}      
\usepackage{xcolor}         

\usepackage{tikz}
\usetikzlibrary{bayesnet}

\usepackage{environ}
\makeatletter
\newsavebox{\measure@tikzpicture}
\NewEnviron{scaletikzpicturetowidth}[1]{%
  \def\tikz@width{#1}%
  \begin{lrbox}{\measure@tikzpicture}%
  \BODY
  \end{lrbox}%
  \pgfmathparse{#1/\wd\measure@tikzpicture}%
  \BODY
}
\makeatother

\usepackage{multirow}
\usepackage{placeins} 
\usepackage{wrapfig} 
\usepackage{amsmath}
\usepackage{amssymb}
\usepackage{subcaption}

\usepackage{algorithm}
\usepackage{algorithmic}

\usepackage{graphicx}
\usepackage{colortbl}
\usepackage{array}

\newcommand{\bs}{\boldsymbol}

\newcommand{\etal}{{\em et al.}}

\renewcommand{\k}{{\mathrm{k}}}

\newcommand{\kbar}{\bar{\k}}
\newcommand{\kdomain}[1]{\k_{\text{#1}}}

\newcommand{\x}{{\bs{x}}}
\renewcommand{\xi}{\x_{i}}

\newcommand{\z}{{\bs{z}}}

\newcommand{\MIM}{{MIM }}
\newcommand{\AMIM}{{A-MIM }}
\newcommand{\cMIM}{{cMIM }}

\newcommand{\modelname}{Contrastive KERMT }

\hypersetup{hidelinks}

\title{Probabilistic Contrastive Pretraining for Multi-task ADME Property Prediction}

\author{%
    Yifan Xue \\
    NVIDIA \\
    Santa Clara, CA 95051 \\
    \texttt{evax@nvidia.com} \\
    \And
    Srimukh Prasad Veccham \\
    NVIDIA \\
    Santa Clara, CA 95051 \\
    \texttt{sveccham@nvidia.com} \\
    \And
    Saee Paliwal \\
    NVIDIA \\
    Santa Clara, CA 95051 \\
    \texttt{saeep@nvidia.com} \\
    \And
    Tyler Shimko \\
    NVIDIA \\
    Santa Clara, CA 95051 \\
    \texttt{tshimko@nvidia.com} \\
    \And
    Micha Livne \\
    NVIDIA \\
    Santa Clara, CA 95051 \\
    \texttt{mlivne@nvidia.com} \\
}

\begin{document}

\maketitle

\begin{abstract}
Accurate prediction of absorption, distribution, metabolism, and excretion (ADME) properties is critical to drug discovery, but remains challenging because ADME endpoints are noisy, interdependent, and often data-limited.
We propose a molecular graph-transformer pretraining framework that combines chemistry-specific self-supervision with contrastive mutual information machine learning (cMIM). Our method encodes molecular graphs into latent variables, reconstructs SMILES strings from the graph-derived latent codes, and augments the contrastive objective with domain-specific self-supervised chemistry tasks. 
Rather than treating these tasks as auxiliary regularizers with separately tuned loss weights, we formulate reconstruction, contrastive discrimination, and chemistry-specific supervision as unit-weighted log-probability factors in a single probabilistic latent-variable objective.
For fine-tuning, we propose a multi-task GNN readout architecture with task-specific multilayer perceptron heads, preserving shared representation learning while mitigating negative transfer and improving the modeling of heterogeneous, nonlinear task relationships.
Across the Biogen, ExpansionRX, and ChEMBL-MT benchmarks, the resulting Contrastive KERMT pretraining strategy consistently improves downstream prediction on these multi-task ADME benchmarks relative to the KERMT baseline with gains of $7.6\%$ on Biogen, $9.9\%$ on ExpansionRX, and $9.5\%$ on ChEMBL-MT datasets averaging over endpoints with statistically significant differences.
We further show that adding ADME-adjacent molecules to the pretraining corpus improves transfer, and that the contrastive component sharpens chemically meaningful latent neighborhoods. These results suggest that cMIM improves ADME representation learning by adding global latent-neighborhood shaping to KERMT's local chemistry-specific self-supervision, and that the combination transfers better than either component alone under controlled pretraining and the stated fine-tuning protocols.
\end{abstract}

\newlength{\mainbodytextfloatsep}
\newlength{\mainbodyfloatsep}
\newlength{\mainbodyintextsep}
\setlength{\mainbodytextfloatsep}{\textfloatsep}
\setlength{\mainbodyfloatsep}{\floatsep}
\setlength{\mainbodyintextsep}{\intextsep}
\setlength{\textfloatsep}{8pt plus 1pt minus 2pt}
\setlength{\floatsep}{6pt plus 1pt minus 1pt}
\setlength{\intextsep}{6pt plus 1pt minus 1pt}

\section{Introduction}
Absorption, distribution, metabolism, and excretion (ADME) properties are central to drug discovery because they determine whether a potent molecule can become a viable therapeutic candidate: it must reach the right exposure, persist for an appropriate duration, and avoid unfavorable pharmacokinetic behavior~\citep{Balani2005,Pellegatti01022012}. In practice, ADME optimization is closely tied to safety assessment, since poor pharmacokinetics can amplify toxicity risk and promising compounds must ultimately balance exposure, efficacy, and safety. Measuring these properties requires a mixture of physicochemical assays, in vitro experiments, and in vivo studies, which are costly, slow, and often available only for limited regions of chemical space. As a result, drug discovery programs increasingly rely on in silico models to prioritize compounds before experimental testing~\citep{lombardo2017,Cáceres01112020,Beckers2023}.

Pretrained molecular graph models are a natural fit for this setting, but practical ADME prediction still poses three coupled challenges. First, learned latent neighborhoods should be chemically meaningful: molecules that are close in representation space should also be close in structure and property space. Second, downstream ADME datasets are multi-task, noisy, and imbalanced: multi-task prediction can regularize data-limited endpoints through shared representations, but a single shared prediction head can underfit endpoint-specific nonlinearities or induce negative transfer when task correlations are weak~\citep{Xu2017Demystifying}. Third, the unlabeled molecules used for pretraining may differ substantially from the ADME assays used for fine-tuning, making corpus design an important part of transfer. Existing molecular GNNs, self-supervised objectives, contrastive methods, and multi-task formulations address parts of this picture; we review them in Section~\ref{sec:related-work}.

We introduce Contrastive KERMT, a graph-transformer pretraining and fine-tuning framework for ADME prediction. The pretraining objective adapts contrastive Mutual Information Machine learning (\cMIM) to molecules by encoding a molecular graph, reconstructing its SMILES representation, and using an in-batch contrastive term to shape the latent space. Our main methodological change is to incorporate chemistry-specific self-supervised tasks, including KERMT/GROVER-style atom, bond, and functional-group prediction \citep{adrian2025multitask}, as observed variables in the same probabilistic latent-variable objective. These tasks are therefore unit-weighted log-probability factors rather than auxiliary regularizers with separately tuned loss weights. For downstream prediction, we combine the pretrained backbone with task-specific MLP heads so each endpoint can learn its own late-stage transformation while still sharing a common molecular representation.

Our contributions are threefold: \textbf{(i)} we propose a probabilistic extension of cMIM that combines global latent-neighborhood shaping with KERMT-style chemistry-specific self-supervision in a single latent-variable objective, avoiding an additional hyperparameter search over auxiliary regularization weights; \textbf{(ii)} we introduce task-specific multi-layer perceptron heads for ADME fine-tuning, where each endpoint-specific head is updated only by its corresponding task loss; and \textbf{(iii)} we show that Contrastive KERMT improves downstream ADME prediction across Biogen, ExpansionRX, and ChEMBL-MT, and that adding ADME-aligned molecules to the pretraining corpus further improves transfer.

\section{Method}

\begingroup
\setlength{\abovedisplayskip}{4pt plus 1pt minus 1pt}
\setlength{\belowdisplayskip}{4pt plus 1pt minus 1pt}
\setlength{\abovedisplayshortskip}{3pt plus 1pt minus 1pt}
\setlength{\belowdisplayshortskip}{3pt plus 1pt minus 1pt}

\paragraph{Graph-to-SMILES Pretraining}\label{sec:mim-loss}
We use contrastive Mutual Information Machine learning~\citep{livne2025contrastive} to shape the latent space of a molecular graph transformer before ADME fine-tuning. For molecule $i$, let $\x_i=(g_i,s_i)$ denote the underlying molecule together with two equivalent molecular views: its 2D molecular graph $g_i$ and its canonical SMILES string $s_i$~\citep{Weininger1988SMILES}. We use $g_i$ and $s_i$ as interchangeable identifiers of the same molecule $\x_i$, while assigning them different roles in the graph-to-SMILES model. The encoder is the KERMT graph-transformer backbone and defines a variational distribution $q_\theta(\z \mid g_i)$ over molecule-level latent codes. The decoder is an autoregressive SMILES transformer and defines $p_\theta(s_i \mid \z_i)$. Thus, unlike a standard graph autoencoder, the model encodes graph representation and reconstructs the molecular string. This graph-to-SMILES construction encourages the latent code to preserve information that is shared across molecular views rather than overfitting to a single input representation. A schematic overview of the \MIM and \cMIM graphical model is provided in Figure~\ref{fig:mim-cmim-overview} in the Appendix.

\paragraph{MIM objective}
The original \MIM objective learns a latent-variable model that maximizes mutual information between inputs and latent codes while encouraging clustered latent structure~\citep{livne2019high,livne2020sentencemim}. Here we use \AMIM, an asymmetric variant where only the posterior is sampled during training, but not the prior. Given a batch $\mathcal{B}=\{\x_i=(g_i,s_i)\}_{i=1}^{B}$ and latent samples $\z_i \sim q_\theta(\z \mid g_i)$, the graph-to-SMILES A-MIM loss is
\begin{equation}
\hat{\mathcal{L}}_{\text{A-MIM}}
= -\frac{1}{B}\sum_{i=1}^{B}
\left[
    \log p_\theta(s_i \mid \z_i)
    + \frac{1}{2}\left(
        \log q_\theta(\z_i \mid g_i) + \log p(\z_i)
    \right)
\right],
\label{eq:amim-graph-smiles}
\end{equation}
where $p(\z)$ is a standard normal prior. The first term rewards reconstruction of the SMILES view from the graph-derived latent code. The symmetric latent-density term keeps sampled codes likely under both the encoder and the prior, which regularizes the representation and encourages a structured latent distribution.
The SMILES likelihood is evaluated autoregressively as $\log p_\theta(s_i \mid \z_i)=\sum_{u=1}^{|s_i|}\log p_\theta(s_{i,u} \mid s_{i,<u},\z_i)$. We use teacher forcing with a character-level SMILES tokenizer and do not length-normalize the reconstruction term.

\paragraph{Contrastive MIM}
\cMIM adds a binary contrastive-like auxiliary random variable that does not require in-batch positive pairs~\citep{livne2025contrastive} and encourages global separation of the latent codes. Instead, it uses the batch to estimate the contribution of mismatched negative samples. We define the batch-conditioned probability
\begin{equation}
p_\theta(\kdomain{con}=1 \mid g_i,\z_i;\mathcal{B})
=
\frac{
    \exp(\operatorname{sim}(\z_i,\z_i)/\tau)
}{
    \exp(\operatorname{sim}(\z_i,\z_i)/\tau)
    + \frac{1}{B-1}\sum_{j \neq i}
        \exp(\operatorname{sim}(\z_i,\z_j)/\tau)
},
\label{eq:cmim-contrastive-prob}
\end{equation}
where $\operatorname{sim}(\cdot,\cdot)$ is cosine similarity and $\tau$ is a temperature. Unlike augmentation-based contrastive objectives, the positive logit in cMIM is not a learned attraction between two augmented molecular views. With cosine similarity, $\operatorname{sim}(\z_i,\z_i)=1$, so the positive score is the fixed reference value $\exp(1/\tau)$. The attraction between a molecule and its latent code is supplied by the MIM terms, namely the graph-to-SMILES reconstruction likelihood and the latent-density terms. The cMIM binary variable instead compares this matched sample against an in-batch Monte Carlo estimate of mismatched latent similarities, thereby repelling unrelated samples and improving global angular separation. Thus, cMIM removes the need for positive-pair molecular augmentations while still using in-batch mismatched samples to estimate the negative expectation. The cMIM loss is then
\begin{equation}
\hat{\mathcal{L}}_{\text{cMIM}}
=
\hat{\mathcal{L}}_{\text{A-MIM}}
- \frac{1}{B}\sum_{i=1}^{B}
\log p_\theta(\kdomain{con}=1 \mid g_i,\z_i;\mathcal{B}).
\label{eq:cmim-loss}
\end{equation}
The reconstruction and latent-density terms preserve local information and cluster structure, while the contrastive term spreads unrelated samples apart in the latent space. This combination is useful for ADME modeling because downstream predictors need both chemically smooth neighborhoods and discriminative separation between molecules with different property profiles.

\paragraph{Chemistry-specific auxiliary variables}
We propose to augment the cMIM loss with domain-specific self-supervised tasks by extending the contrastive auxiliary variable with chemistry-specific variables. This turns the contrastive objective from a purely instance-discrimination signal into a joint objective where latent codes also predict chemically meaningful structure. Let $y_{i,t}=\operatorname{task}_t(g_i,s_i)$ be the target for chemistry task $t$, computed without assay labels, and let $p_\theta(\kdomain{t}=y_{i,t}\mid g_i,\z_i)$ be the corresponding prediction head. In our setting these tasks are the chemically informed KERMT \citep{adrian2025multitask} pretraining targets, such as atom-context, bond-context, and functional group prediction, when those heads are active. Defining $\kbar=(\kdomain{con},\kdomain{1},\ldots,\kdomain{T})$, the extended objective is
\begin{equation}
\hat{\mathcal{L}}_{\text{x-cMIM}}
= \hat{\mathcal{L}}_{\text{cMIM}}
- \frac{1}{B}\sum_{i,t}
\log p_\theta(\kdomain{t}=y_{i,t}\mid g_i,\z_i).
\label{eq:extended-cmim-loss}
\end{equation}
Equation~\eqref{eq:extended-cmim-loss} sums over batch elements and active chemistry tasks and reduces to Eq.~\eqref{eq:cmim-loss} when $T=0$; its expanded log-probability form is given in Appendix~\ref{app:cmim}. Written this way, reconstruction, latent regularization, contrastive discrimination, and chemistry-specific supervision are all log-probability factors in a single joint objective. By ``unit weighting'' we mean unit scalar coefficients on each log-probability factor --- not equal gradient magnitudes. Each factor enters as its native sample-level log-probability: the SMILES likelihood summed over tokens, latent-density terms over latent dimensions, and chemistry objectives reduced as in KERMT. Algorithm~\ref{algo:cmim} in Appendix~\ref{app:cmim} summarizes the full pretraining loop.
\endgroup

\section{Experimental Setup}

\subsection{Datasets and Benchmarks}\label{sec:datasets}

We use datasets in two distinct roles. \emph{Pretraining corpora} provide unlabeled molecules for learning the encoder representations; these corpora vary in scale and in-domain ADME coverage. \emph{Downstream benchmarks} provide labeled assay endpoints for evaluating property prediction after fine-tuning. 
The pretraining corpora are ZINC15\_ChEMBL-11M, the larger ZINC15\_ChEMBL-up-208M corpus, and several ADME-adjacent augmentations built from Biogen molecules, MolMIM-generated molecules seeded from Biogen, ExpansionRX, and ChEMBL-MT. The downstream benchmarks are Biogen, ExpansionRX, and ChEMBL-MT, containing 4, 9, and 25 prediction tasks respectively. Detailed information about dataset construction, curation, and split definitions is provided in Appendix~\ref{app:datasets}.

\subsection{Pretraining Model Variants}\label{sec:model-variants}

Within the extended cMIM framework of Eq.~\eqref{eq:extended-cmim-loss}, we compare three pretraining variants that toggle two components: the cMIM block (graph-to-SMILES reconstruction plus in-batch contrastive discrimination) and the KERMT chemistry-specific pretraining targets (atom-context, bond-context, and functional group prediction). All variants share the KERMT (Kinetic GROVER Multi-Task)~\citep{adrian2025multitask} encoder, an extension of the GROVER graph transformer~\citep{rong2020self} with local atom/bond message passing, global attention, and four encoder readouts available for downstream use: two atom-level and two bond-level readouts obtained by pooling atom and bond features under atom-context and bond-context attention. The encoder uses hidden size 800, 6 message-passing-plus-attention layers, 4 attention heads per layer, 1 MT block, PReLU activations, and 0.1 dropout. Component-level differences are summarized in Table~\ref{tab:model-variants}.

\paragraph{KERMT} This is the cMIM-block-off ablation and recovers the original KERMT pretraining recipe: the encoder is trained only with atom-context prediction, bond-context prediction, and functional group classification heads applied from both atom and bond representations.

\paragraph{cMIM-only} This variant removes the KERMT pretraining-target heads and trains the encoder with the base cMIM objective alone---Eq.~\eqref{eq:cmim-loss}, equivalently Eq.~\eqref{eq:extended-cmim-loss} with $T=0$---so its only pretraining signals are graph-to-SMILES reconstruction and in-batch contrastive discrimination. A mean-pooled encoder readout is mapped through a learned projection to a 512-dimensional latent distribution, and a 3-layer, 8-head transformer decoder~\citep{Vaswani2017Attention} with 2048 FFN dimensions and rotary positional encoding~\citep{Su2024RoFormer} reconstructs the input SMILES from the latent code.

\begin{figure}[t]
    \centering
    \begin{tikzpicture}[
    font=\scriptsize,
    box/.style={
        draw,
        rounded corners=1.5pt,
        align=center,
        inner xsep=4pt,
        inner ysep=3pt,
        minimum height=0.55cm
    },
    core/.style={box, fill=blue!8, minimum width=1.65cm},
    view/.style={box, fill=gray!10, minimum width=1.2cm},
    head/.style={box, fill=green!8, minimum width=1.45cm},
    cmim/.style={box, fill=orange!10, minimum width=1.45cm},
    loss/.style={box, fill=red!7, minimum width=1.35cm},
    molecule/.style={
        draw,
        rounded corners=2pt,
        align=center,
        minimum width=1.55cm,
        minimum height=1.55cm
    },
    arrow/.style={->, line width=0.4pt}
]
    \node[molecule, label={[font=\scriptsize]above:$\x_i$}] (mol) at (0.0,0.0) {};
    \node[view] (graph) at (0.0,0.36) {$g_i$\\graph};
    \node[view] (smiles) at (0.0,-0.36) {$s_i$\\SMILES};
    \node[core] (enc) at (2.25,0.0) {KERMT\\encoder};

    \node[head] (vocab) at (4.85,0.72) {atom/bond\\vocab};
    \node[head] (fg) at (6.95,0.72) {functional\\groups};
    \node[loss] (vloss) at (9.05,0.72) {$\mathcal{L}_{\mathrm{vocab}}$};

    \node[cmim] (latent) at (4.85,-0.72) {latent\\$\z_i$};
    \node[cmim] (decoder) at (6.95,-0.72) {Transformer\\decoder};
    \node[cmim] (closs) at (9.05,-0.72) {$\mathcal{L}_{\mathrm{rec}}$\\$+\mathcal{L}_{\mathrm{con}}$};

    \node[loss] (hybrid) at (12.0,0.0) {Hybrid\\objective};

    \draw[arrow] (graph) -- (enc);
    \draw[arrow] (enc) -- (vocab);
    \draw[arrow] (vocab) -- (fg);
    \draw[arrow] (fg) -- (vloss);
    \draw[arrow] (enc) -- (latent);
    \draw[arrow] (latent) -- (decoder);
    \draw[arrow] (decoder) -- (closs);
    \draw[arrow, rounded corners=2pt] (latent.north) -- ++(0,0.28) -| (closs.north);
    \coordinate (smilesRoute) at (0.0,-1.35);
    \coordinate (lossRoute) at (9.05,-1.35);
    \draw[arrow, rounded corners=2pt] (smiles.south) -- (smilesRoute) -- (lossRoute) -- (closs.south);
    \draw[arrow] (vloss) -- (hybrid);
    \draw[arrow] (closs) -- (hybrid);

    \node[align=center] at (6.95,1.35) {\textbf{KERMT path}};
    \node[align=center] at (6.95,-1.85) {\textbf{cMIM path}};
\end{tikzpicture}
    \caption{Architecture of Contrastive KERMT. KERMT uses the shared graph-transformer encoder with vocabulary-prediction heads; cMIM-only uses the encoder with the SMILES decoder and cMIM loss; Contrastive KERMT combines both paths. See Table~\ref{tab:model-variants} for a component-level comparison.}
    \label{fig:architectures}
\end{figure}
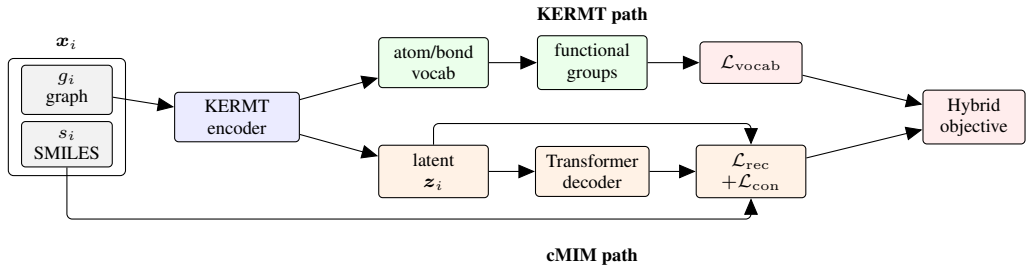

\paragraph{Contrastive KERMT} \modelname activates both cMIM and the KERMT pretraining targets, yielding the full objective in Eq.~\eqref{eq:extended-cmim-loss}. The auxiliary targets are treated as additional observed variables in the joint latent-variable model rather than as separately weighted regularizers, so the local atom/bond supervision and the global contrastive-reconstruction signal enter as unit-weighted log-probability factors. This preserves the probabilistic interpretation and avoids loss-weight tuning. The architecture therefore includes all components of KERMT and cMIM-only (Figure~\ref{fig:architectures}). Pretraining hyperparameters and implementation details for all three variants are reported in Appendix~\ref{app:implementation-details}.

\subsection{Pretraining Corpora and Configurations}\label{sec:experiments}
Each pretraining configuration is a pairing of one model variant from Section~\ref{sec:model-variants} with one unlabeled molecular corpus from Section~\ref{sec:datasets}. Table~\ref{tab:pretraining-corpora} lists the configurations evaluated in this work. The first three rows isolate the effect of the pretraining objective at fixed corpus scale by training KERMT, cMIM-only, and Contrastive KERMT on the same 11M-molecule base corpus. The remaining rows keep the full Contrastive KERMT objective fixed and vary the pretraining corpus.

The corpus ablations span a size axis (11M $\to$ 208M molecules) and three in-domain augmentation strategies layered on top of the base corpus: Biogen molecules alone; Biogen plus 30k synthetic molecules generated from Biogen seeds; and a pooled ADME-adjacent corpus combining Biogen, ExpansionRX, and ChEMBL-MT. We use MolMIM~\citep{reidenbach2022improving} to produce the synthetic extensions, and verify in Appendix~\ref{app:embedding-viz} that they remain close to their Biogen seeds in chemical space. Implementation details, including training budgets, are reported in Appendix~\ref{app:implementation-details}.
Some in-domain pretraining configurations include molecules that also appear in downstream validation or test splits, always without assay labels. These settings therefore evaluate label-free corpus adaptation rather than a strict molecule-holdout pretraining protocol. Supervised fine-tuning and model selection still use the downstream train/validation/test labels only according to the stated splits.

\begin{table}[ht]
    \centering
    \small
    \caption{Pretraining configurations. See full corpus definitions in Appendix~\ref{app:datasets}.}
    \label{tab:pretraining-corpora}
    \setlength{\tabcolsep}{3pt}
    \begin{tabular}{l>{\raggedright\arraybackslash}p{0.62\linewidth}r}
        \toprule
        Model Variant & Pretraining corpus & Samples \\
        \midrule
        KERMT             & ZINC15\_ChEMBL-11M & 11M \\
        cMIM-only         & ZINC15\_ChEMBL-11M & 11M \\
        Contrastive KERMT & ZINC15\_ChEMBL-11M & 11M \\
        Contrastive KERMT & ZINC15\_ChEMBL-up-208M & 208M \\
        \midrule
        Contrastive KERMT & ZINC15\_ChEMBL-11M + Biogen & 11.004M \\
        Contrastive KERMT & ZINC15\_ChEMBL-up-208M + Biogen & 208.004M \\
        Contrastive KERMT & ZINC15\_ChEMBL-11M + Biogen + 30k generated & 11.034M \\
        Contrastive KERMT & ZINC15\_ChEMBL-11M + Biogen + ExpansionRX + ChEMBL-MT & 11.125M \\
        \bottomrule
    \end{tabular}
\end{table}

\subsection{Downstream Fine-tuning Protocol}\label{sec:downstream}

For downstream evaluation, each pretraining configuration produces an encoder initialization that is fine-tuned end-to-end on each labeled ADME benchmark. The downstream tasks are the assay endpoints within a benchmark: 4 tasks for Biogen, 9 for ExpansionRX, and 25 for ChEMBL-MT. We compare a task-specific readout against a default shared-head readout to distinguish downstream prediction architecture from pretraining variant or corpus; all fine-tuning implementation details are provided in Appendix~\ref{app:implementation-details}. A complementary representation-quality assessment via linear probing on frozen embeddings is provided in Appendix~\ref{app:probing}.
All pretraining-only modules, including the SMILES decoder, posterior parameter head, and cMIM contrastive branch, are discarded before downstream fine-tuning. Every pretrained variant is fine-tuned through the same KERMT encoder/readout interface, so downstream differences reflect the learned encoder initialization rather than additional cMIM inference-time capacity.

\section{Results}
For the results shown below, we use different metrics for different experimental questions. Section~\ref{sec:results-contrastive-effects} evaluates representation geometry, so we use nearest-neighbor preservation metrics: Trustworthiness@$10$, Precision@$10$, Violation Rate, and conditional $Q_{0.9}$ distance. Section~\ref{sec:comp_with_baselines} compares predictive performance across model families using Pearson's $R$, which is scale-invariant and standard for comparing endpoint-wise ranking and correlation across heterogeneous ADME assays. Section~\ref{sec:results-finetune} analyzes within-family pretraining and corpus ablations using MAE, because MAE measures absolute assay-prediction error and is directly interpretable within each benchmark. Full fine-tuning runs provide the main downstream evidence, while latent-neighborhood preservation and frozen linear probing are diagnostic representation-level analyses used to interpret how cMIM changes the encoder. Metric definitions and endpoint-level results are provided in the appendices.

\subsection{Effect of cMIM on Latent Neighborhoods}\label{sec:results-contrastive-effects}

To test whether cMIM improves latent geometry, we compare the cMIM-only encoder with the matched KERMT-style baseline on Biogen and ExpansionRX. For each molecule, we retrieve $k=10$ latent nearest neighbors and evaluate whether those neighbors remain close along three chemically meaningful axes: canonical-SMILES edit distance, Morgan-fingerprint distance~\citep{Rogers2010Extended} ($1-$Tanimoto similarity~\citep{Bajusz2015Tanimoto}), and measured ADME-property distance. We report Trustworthiness@$10$~\citep{Venna2001Neighborhood} and Precision@$10$ (higher is better), plus Violation Rate and conditional $Q_{0.9}$ distance (lower is better); Appendix~\ref{app:latent-geometry-metrics} gives the metric definitions.

\begin{table}[ht]
  \centering
  \small
  \caption{cMIM-normalized relative advantage ($\Delta\%$) over the matched KERMT baseline on latent-neighborhood preservation for Biogen (\textbf{B}) and ExpansionRX (\textbf{E}). Columns are the comparison spaces used to judge latent-neighbor quality. For higher-is-better metrics, $\Delta\%=100\cdot(m_{\mathrm{cMIM}}-m_{\mathrm{KERMT}})/|m_{\mathrm{cMIM}}|$. For lower-is-better metrics we switch the subtraction order. Positive values always indicate that cMIM improves over the matched KERMT baseline.}
  \begin{tabular}{l cc cc cc}
    \toprule
    Metric
      & \multicolumn{2}{c}{\textbf{SMILES}}
      & \multicolumn{2}{c}{\textbf{Morgan FP}}
      & \multicolumn{2}{c}{\textbf{Property}} \\
    \cmidrule(lr){2-3}\cmidrule(lr){4-5}\cmidrule(lr){6-7}
      & \textbf{B} & \textbf{E}
      & \textbf{B} & \textbf{E}
      & \textbf{B} & \textbf{E} \\
    \midrule
    $\uparrow$ Trustworthiness@$10$ & 10.7\% &  9.2\% &  3.0\% &  3.3\% &  1.1\% &  2.0\% \\
    $\uparrow$ Precision@$10$       & 40.3\% & 30.1\% & 42.3\% & 28.3\% &  0.0\% & 10.0\% \\
    $\downarrow$ Violation Rate     & 13.1\% & 24.4\% & 27.6\% & 22.7\% &  0.2\% &  2.0\% \\
    $\downarrow$ Cond.\ $Q_{0.9}$   &  6.5\% & 10.8\% &  2.0\% & 17.6\% &  2.4\% &  0.8\% \\
    \bottomrule
  \end{tabular}
  \label{tab:latent_property_delta}
\end{table}

Table~\ref{tab:latent_property_delta} shows that cMIM improves nearly every metric, comparison axis, and benchmark. In cMIM-normalized relative terms, the largest gains are in neighborhood overlap: Precision@$10$ rises by $40.3\%$/$42.3\%$ on Biogen for SMILES/Morgan FP and by $30.1\%$/$28.3\%$ on ExpansionRX, with a further $10.0\%$ gain in ExpansionRX property space. Failure-oriented metrics move in the same direction: Violation Rate drops most on Biogen Morgan FP ($27.6\%$) and ExpansionRX SMILES ($24.4\%$), and conditional $Q_{0.9}$ distance decreases for all axes on both benchmarks. Thus, the contrastive objective primarily sharpens chemically local neighborhoods while preserving property-relevant structure.

\subsection{Performance on ADME Benchmarks}\label{sec:comp_with_baselines}
\begin{table}[ht]
    \centering
    \small
    \caption{Number of ADME tasks on which each model is the best or statistically indistinguishable from the best under Tukey HSD, using Pearson's R metric across Biogen, ExpansionRX, and ChEMBL-MT datasets. Bold marks the per-dataset maximum in the ``Best'' column. Fixed descriptor baselines are excluded here because they underperform the neural baselines in this multi-task setting.}
    \begin{tabular}{lcccccc}
    \toprule
    & \multicolumn{2}{c}{Biogen}
    & \multicolumn{2}{c}{ExpansionRX}
    & \multicolumn{2}{c}{ChEMBL-MT} \\
    \cmidrule(lr){2-3} \cmidrule(lr){4-5} \cmidrule(lr){6-7}
                     Model    & Best    & Indist. & Best      & Indist.  & Best     & Indist.  \\
                     \midrule
\modelname & \textbf{3}       & 1                & \textbf{5}         & 4                   & \textbf{14}       & 10                 \\
KERMT                                    & 0       & 4                & 2         & 5                   & 1        & 22                 \\
KPGT                                     & 0       & 3                & 0         & 5                   & 5        & 14                 \\
MolCLR                                   & 0       & 1                & 0         & 1                   & 1        & 7                  \\
Chemprop                                 & 1       & 3                & 2         & 3                   & 1        & 8   \\
    \bottomrule
    \end{tabular}
    \label{tab:tukey-wins-tie-count}
\end{table}

We compare the performance of \modelname against both fixed descriptors and graph neural network models, including both pretrained and non-pretrained variants. We evaluate on three ADME-focused multi-task datasets: Biogen, ExpansionRX, and ChEMBL-MT. All neural baselines are evaluated under the same downstream split, preprocessing, target transformations, missing-label handling, early-stopping criterion, and benchmark-specific evaluation metric. Pretraining data access differs across model families. Within-family KERMT comparisons use matched pretraining-corpus access when isolating the effect of adding cMIM. The pretraining molecular corpus of KERMT and KPGT (Knowledge-guided Pretraining Graph Transformer) models overlap with ChEMBL-MT. By contrast, ExpansionRX is new and was not included in the pretraining corpus of external baselines. Chemprop, Morgan+MLP, and RDKit+MLP use no unsupervised pretraining. We therefore interpret cross-baseline results as comparisons of complete modeling recipes under the same downstream evaluation setup, while within-family KERMT ablations isolate objective and corpus effects under controlled pretraining access. Although MoleculeNet~\citep{Wu2018MoleculeNet} and Therapeutic Data Commons~\citep{huang2022artificial} datasets have been used extensively in prior work, we do not benchmark on them due to data curation errors, invalid structures, and limited relevance to real-world multi-objective drug discovery process tasks~\citep{walters2023benchmarks}. See Appendix~\ref{app:datasets} for detailed dataset descriptions.

\begin{figure}[!ht]
    \centering
    \includegraphics[width=0.9\linewidth]{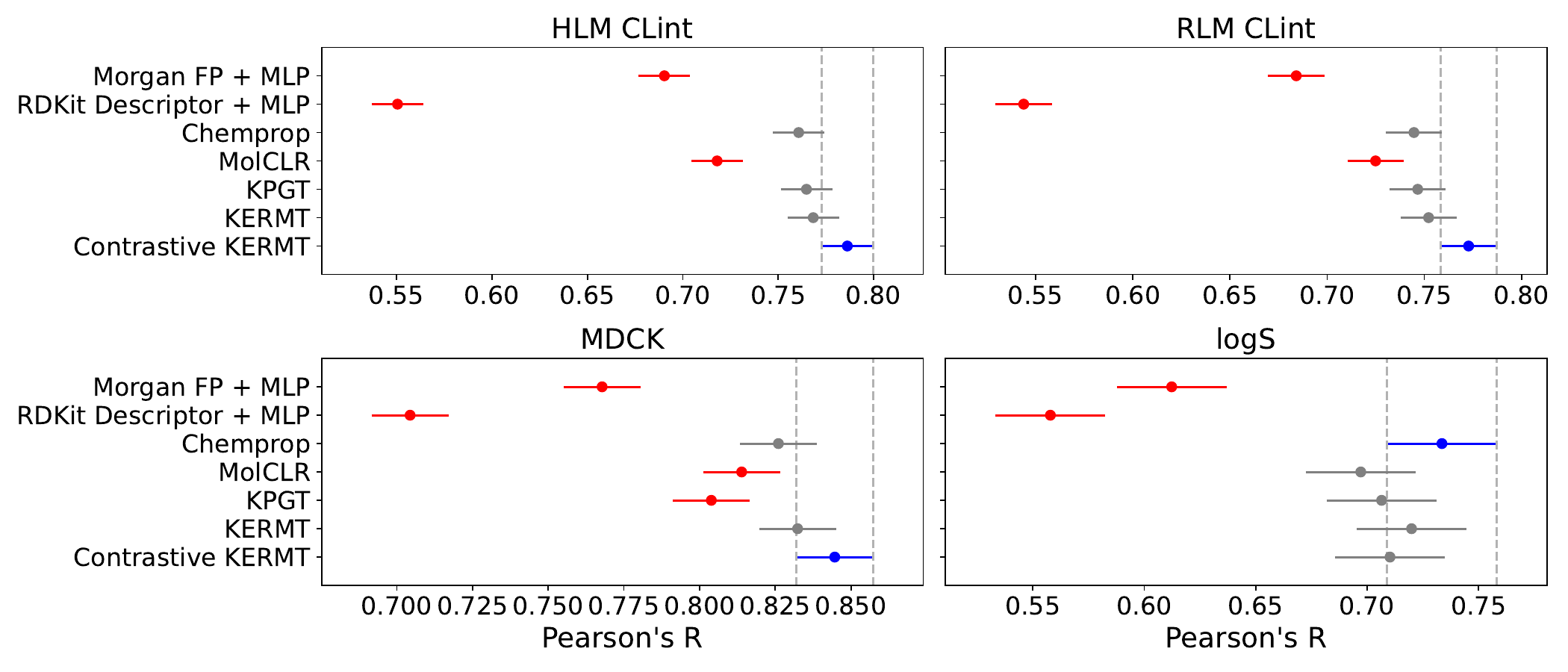}
    \caption{Tukey HSD plot of Pearson's R comparing \modelname model performance on Biogen dataset with all baselines. ANOVA p-values $<10^{-4}$ for all assays. Blue: best method, Gray: statistically indistinguishable from the best method, Red: statistically worse than the best method.}
    \label{fig:comp_with_baselines-biogen}
\end{figure}

In Table~\ref{tab:tukey-wins-tie-count}, \modelname has the best performance on 3 of 4 endpoints in the Biogen dataset, 5 of 9 endpoints in the ExpansionRX dataset, and 14 out of 25 endpoints in the ChEMBL-MT dataset. When \modelname does not have the best performance, it is usually statistically indistinguishable from the best under Tukey HSD. For example, in the ChEMBL-MT dataset, our \modelname model almost always shows the best or statistically indistinguishable from the best performance across endpoints (see Figure~\ref{fig:comp_with_baselines-chembl-mt}). KPGT~\citep{Li2023}, pretrained on RDKit fingerprint and molecular descriptors, shows the next best performance. The KERMT model also shows strong performance, being statistically indistinguishable from the best-performing model on 22 ADME endpoints. Chemprop, a widely used GNN-based molecular property prediction model based on the directed message passing neural network architecture~\citep{Yang2019analyzing,Heid2024,graff2025}, is not a pretrained model and shows the best performance on permeability assays in the ExpansionRX dataset (see Figure~\ref{fig:comp_with_baselines-exprx}). In single-source datasets like Biogen (Figure~\ref{fig:comp_with_baselines-biogen}) and ExpansionRX, \modelname achieves the best or statistically indistinguishable performance on most endpoints. Fixed descriptor models underperform the neural baselines in this multi-task setting.

\subsection{Ablation: Which design choices explain the gains?}\label{sec:results-finetune}

To complement the cross-baseline comparison of Section~\ref{sec:comp_with_baselines}, we report a within-family ablation across the pretraining objective (vocab, cMIM, vocab+cMIM), the corpus size (11M $\to$ 208M molecules), and three flavors of in-domain augmentation (Biogen seeds, MolMIM-generated extensions, and pooled ADME-adjacent corpora). Each pretrained backbone is fine-tuned on Biogen, ExpansionRX, and ChEMBL-MT following the protocol of Section~\ref{sec:downstream}.

\paragraph{cMIM complements KERMT but does not replace it.}
Adding the cMIM objective on top of the original KERMT vocabulary-prediction signals yields significant-endpoint conditional improvements (mean per-endpoint $\Delta\%$ over endpoints with a significant two-sample $t$-test vs.\ \texttt{kermt\_base}; Appendix~\ref{app:finetune-significance}) on every benchmark (Biogen $+7.6\%$, ExpansionRX $+8.7\%$, ChEMBL-MT $+8.7\%$). cMIM-only, however, is statistically indistinguishable from the baseline on Biogen and ChEMBL-MT, and significantly \emph{worse} than baseline on ExpansionRX. The vocabulary-prediction signal therefore remains essential, and the observed gains come from combining the two objectives rather than from the contrastive objective in isolation.
We do not evaluate a contrastive-only variant because the cMIM contrastive behavior is defined as part of the complete MIM objective rather than as a standalone loss~\citep{livne2025contrastive}.

\begin{figure}[ht]
    \centering
    \includegraphics[width=\linewidth]{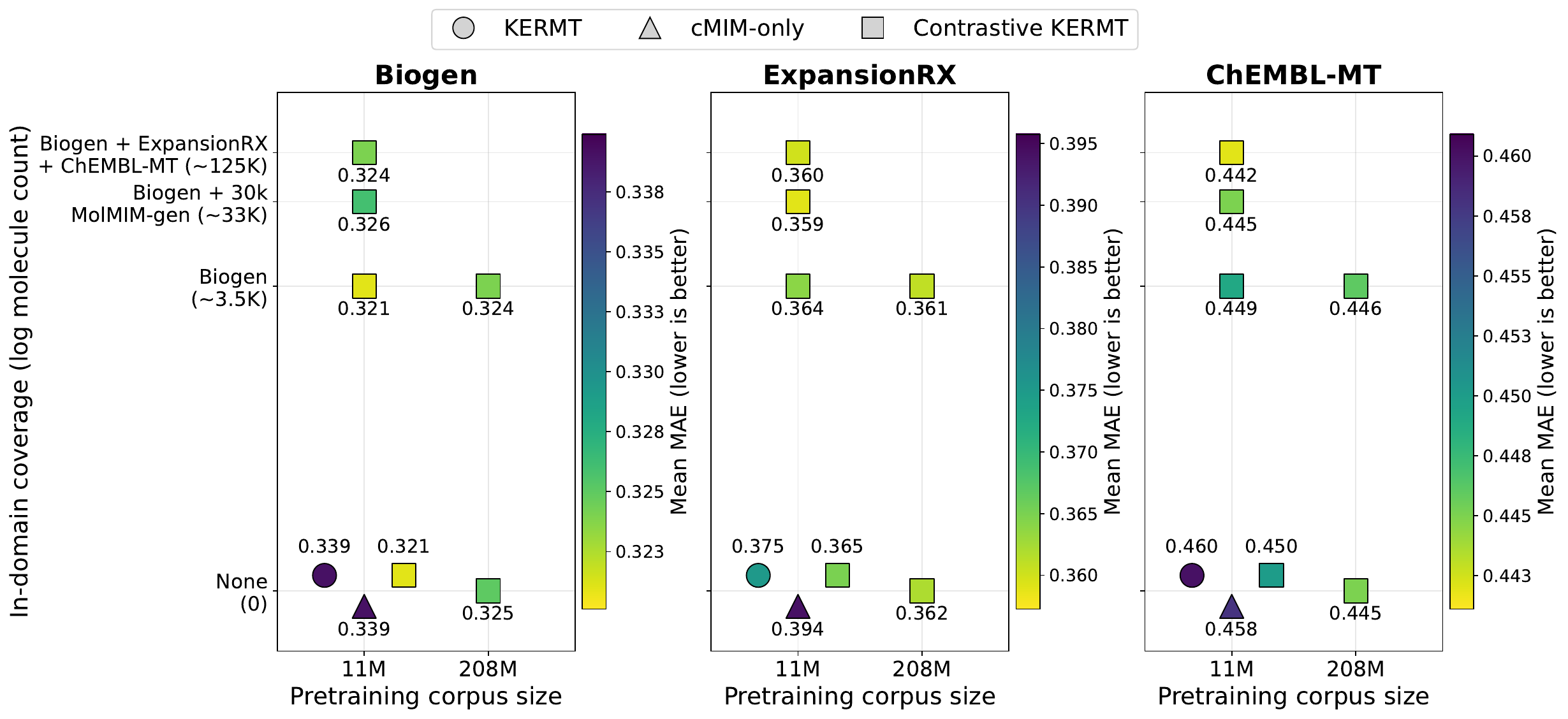}
    \caption{Effect of pretraining-corpus scale and in-domain ADME coverage on downstream mean MAE. The horizontal axis indicates base corpus size; the vertical axis indicates added ADME-adjacent molecules on a log-count scale. Marker shape denotes pretraining objective; color and printed value denote mean MAE. Only evaluated configurations are shown.}
    \label{fig:aug-axes-finetune}
\end{figure}

\paragraph{Corpus expansion improves performance, with pooled ADME-adjacent corpora generalizing most consistently.}Figure~\ref{fig:aug-axes-finetune} visualizes how mean MAE varies across the two pretraining-corpus expansion axes (size and in-domain coverage) for each benchmark.
Scaling the base corpus from 11M to 208M improves ExpansionRX and ChEMBL-MT under mean MAE, but not Biogen; Biogen benefits more from in-domain Biogen augmentation than from broad corpus scaling alone. Consistent with this, adding in-domain Biogen molecules to the 11M base matches the best Biogen mean MAE (Table~\ref{tab:finetune-improvement}) and substantially improves ExpansionRX transfer; further mixing in 30k MolMIM-generated extensions gives the strongest configuration on ExpansionRX ($+9.9\%$), suggesting that exposure to Biogen-seeded generative neighborhoods transfers to a held-out ADME dataset. Pooling all three ADME-adjacent corpora (Biogen + ExpansionRX + ChEMBL-MT) is the single best model on ChEMBL-MT ($+9.5\%$, $p < 10^{-4}$) and has the highest worst-case improvement across the three benchmarks (every benchmark sees at least $+6.9\%$), indicating that broader in-domain diversity at pretraining provides the most uniform transfer signal across ADME endpoints.
A complementary frozen-representation diagnostic in Appendix~\ref{app:probing} shows the same pattern: Contrastive KERMT improves average linear-probe performance over KERMT on both Biogen-source and ExpansionRX-source embeddings, whereas cMIM-only is weak. This supports the interpretation that the combined objective changes the pretrained representation itself, not only the fine-tuned prediction head.

\paragraph{Task-specific fine-tuning layers consistently improve performance.}
Across all three benchmarks, matched default-vs-task-specific comparisons for the \texttt{kermt\_base} baseline show that task-specific layers reduce mean MAE: Biogen ($0.339 \to 0.332$), ExpansionRX ($0.380 \to 0.375$), and ChEMBL-MT ($0.466 \to 0.460$). The same pattern holds for Contrastive KERMT on Biogen, where both protocols were run, with reductions of $0.6$--$1.9\%$ across all configurations (Table~\ref{tab:finetune-improvement}).

\section{Related Work}\label{sec:related-work}

\paragraph{Molecular GNNs}
Graph neural networks naturally represent molecules by mapping atoms and bonds to graph structure. Message passing neural networks unified early molecular graph models~\citep{Gilmer2017}, directed message passing remains widely used through Chemprop~\citep{Yang2019analyzing,Heid2024,graff2025}, and attention-based GNNs such as Attentive FP capture task-relevant nonlocal interactions~\citep{Xiong2020}. More recent graph transformers and hybrid message-passing/attention models, including MolGPS, show that molecular GNNs can scale to larger models and datasets~\citep{masters2022gps++,sypetkowski2024scalability}. We build on this 2D graph-transformer lineage and focus on strengthening the pretraining signal for ADME prediction.

\paragraph{Self-supervised molecular pretraining}
Self-supervised learning pretrains on large unlabeled chemical libraries before fine-tuning, and has been shown to improve performance, particularly in low-data regimes.
SMILES Transformer, ChemBERTa, and MoLFormer adapt language-modeling objectives to SMILES strings~\citep{Honda2019SMILESTransformer,Chithrananda2020ChemBERTa,Ross2022MolFormer}, whereas graph and graph-transformer methods pretrain on topology, functional groups, or auxiliary computed chemical properties~\citep{Wan2025}. KPGT injects descriptors and fingerprints~\citep{Li2023} into pretraining. Uni-Mol and SCAGE add 3D conformer information~\citep{Zhou2023UniMol,Qiao2025SCAGE}. Most relevant to our setting, GROVER uses contextual prediction at atom- and bond-level and functional-group identification at molecule level~\citep{rong2020self}; KERMT extends GROVER to multi-task fine-tuning~\citep{adrian2025multitask}. \modelname retains KERMT's atom, bond, and functional-group objectives while adding SMILES reconstruction and the cMIM contrastive objective, allowing us to test whether global latent-space structure complements local chemical prediction tasks.

\paragraph{Contrastive molecular representation learning}
Contrastive learning is a commonly used method in graph and molecular representation learning: InfoGraph maximizes mutual information between graph- and substructure-level representations~\citep{sun2020infograph}, GraphCL defines positives through graph augmentations~\citep{you2020graph}, and MolCLR adapts this recipe to molecular transfer benchmarks~\citep{Wang2022}. These methods rely on augmentation-defined positive pairs. By contrast, MIM learns latent variables by maximizing mutual information while encouraging clustered structure~\citep{livne2019high,livne2020sentencemim}, and cMIM adds in-batch contrastive discrimination without requiring graph perturbations~\citep{livne2025contrastive}. We extend cMIM with chemistry-specific auxiliary variables and KERMT-style vocabulary prediction to preserve local chemical structure while shaping a globally discriminative latent space.

\paragraph{Multi-task ADME modeling}
ADME optimization is multi-objective, and multi-task neural networks have long been used in QSAR and ADME modeling to reflect this~\citep{Dahl2014MultitaskQSAR}. Industry studies show that shared graph representations can help data-sparse endpoints borrow signal from data-rich and/or correlated assays~\citep{Montanari2020Modeling,Walter2024}. Pretrained chemical models have also been adapted to multi-task ADME prediction~\citep{adrian2025multitask}. As multi-task learning can also suffer from task competition or negative transfer when endpoints differ in relation, noise, scale, or protocol~\citep{standley2020tasks}, we use task-specific readout heads on top of shared molecular representations to mitigate this issue.

\section{Discussion}
\modelname combines graph-to-SMILES cMIM with KERMT's atom/bond vocabulary and functional  group prediction tasks as unit-weighted log-probability factors in a single joint objective. The within-family ablations show that these signals are complementary: cMIM-only is weak downstream, while Contrastive KERMT improves over KERMT on all three benchmarks. Latent-neighborhood analysis supports the mechanism, showing improved preservation of SMILES, Morgan fingerprint, and property-neighborhood structure. Adding in-domain ADME molecule to pretraining further improves transfer. Unlike knowledge-guided approaches such as KPGT~\citep{Li2023}, which explicitly predict fingerprints and descriptors, cMIM implicitly incorporates fingerprint-awareness without explicitly using it in pretraining.

\paragraph{Limitations.}\label{sec:limitations}
Our evaluation is restricted to ADME endpoints in three multi-task datasets and one KERMT graph-transformer backbone architecture; transfer to other molecular properties or GNN architectures remains untested. Several corpus-adaptation experiments also include unlabeled downstream validation/test molecules. This reflects realistic label-free chemical space adaptation, but not a strict inductive benchmark, so we interpret these results as label-free/transductive pretraining gains. Industry deployment would also require retraining as target chemical space evolves, adding compute burden for large-scale pretraining~\citep{Rich2024}.

Overall, the evidence supports \modelname as an additive pretraining strategy for ADME representation learning: KERMT supplies local chemistry-specific self-supervision, cMIM reshapes global latent neighborhoods through reconstruction and contrastive separation, and the combination transfers better than either component alone under matched pretraining and fine-tuning protocols.

\setlength{\textfloatsep}{\mainbodytextfloatsep}
\setlength{\floatsep}{\mainbodyfloatsep}
\setlength{\intextsep}{\mainbodyintextsep}

\bibliography{paper}
\bibliographystyle{plainnat}


\appendix
\section{\cMIM Overview}\label{app:cmim}

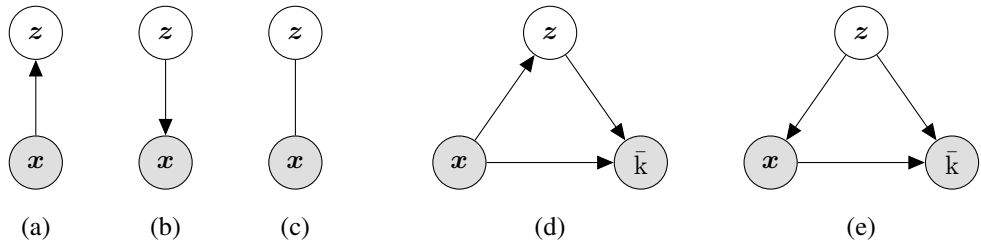
\begin{figure}[ht]
    \centering
    \begin{subfigure}[t]{0.48\textwidth}
        \centering
        \begin{tikzpicture}

    
    \node[latent]                (zenc) {$\z$};
    \node[obs, below=of zenc]                (xenc) {$\x$};
    
    \node[latent, right=of zenc]               (zdec) {$\z$};
    \node[obs, below=of zdec]                (xdec) {$\x$};
    
    \node[latent, right=of zdec]             (z) {$\z$};
    \node[obs, below=of z]                   (x) {$\x$};
    
     \node[const, below=of xenc, yshift=0.65cm]  {(a)} ; %
     \node[const, below=of xdec, yshift=0.65cm]  {(b)} ; %
     \node[const, below=of x, yshift=0.65cm]  {(c)} ; %
    
    \edge [-] {x} {z} ; %
    \edge [bend left] {xenc} {zenc} ; %
    \edge [bend left] {zdec} {xdec} ; %
    
\end{tikzpicture}
        \phantomcaption
        \label{fig:mim-model}
    \end{subfigure}
    \begin{subfigure}[t]{0.48\textwidth}
        \centering
        \begin{tikzpicture}

    
    \node[latent]                                           (zenc) {$\z$};
    \node[obs, below=of zenc, xshift=-1.2cm]                (xenc) {$\x$};
    \node[obs, below=of zenc, xshift=1.2cm]                 (kenc) {$\kbar$};
    
    \node[latent, right=of zenc, xshift=2.4cm]              (zdec) {$\z$};
    \node[obs, below=of zdec, xshift=-1.2cm]                (xdec) {$\x$};
    \node[obs, below=of zdec, xshift=1.2cm]                 (kdec) {$\kbar$};
        
     \node[const, below=of xenc, yshift=0.65cm, xshift=1.2cm]  {(d)} ; %
     \node[const, below=of xdec, yshift=0.65cm, xshift=1.2cm]  {(e)} ; %
    
    \edge  {xenc} {zenc} ; %
    \edge  {zenc} {kenc} ; %
    \edge  {xenc} {kenc} ; %

    \edge  {zdec} {xdec} ; %
    \edge  {zdec}  {kdec} ; %
    \edge  {xdec} {kdec} ; %
    
\end{tikzpicture}
        \phantomcaption
        \label{fig:cmim-model}
    \end{subfigure}
    \caption{(Left) The \MIM view learns compatible encoder and decoder factorizations through a shared latent code. In our molecular instantiation, the encoder consumes a 2D molecular graph and the decoder reconstructs a SMILES sequence. (Right) \cMIM augments this latent-variable objective with auxiliary variables $\kbar$, including a contrastive variable and optional chemistry-specific self-supervised variables.}
    \label{fig:mim-cmim-overview}
\end{figure}

Figure~\ref{fig:mim-cmim-overview} summarizes the latent-variable view used in Section~\ref{sec:mim-loss}. The left panel shows the base \MIM construction, where an encoder maps an observed molecule to a latent code and a decoder reconstructs an observed molecular view from that code. The right panel shows the \cMIM extension used in this work: the same latent code is also connected to auxiliary variables $\kbar$, allowing contrastive discrimination and chemistry-specific self-supervised targets to enter the objective as probabilistic factors rather than separately weighted regularizers.

Expanding Eq.~\eqref{eq:extended-cmim-loss} gives the per-batch objective
\begin{equation}
\begin{aligned}
\hat{\mathcal{L}}_{\text{x-cMIM}}
= -\frac{1}{B}\sum_{i=1}^{B}
\Big[
& \log p_\theta(s_i \mid \z_i)
  + \frac{1}{2}\left(
        \log q_\theta(\z_i \mid g_i) + \log p(\z_i)
    \right) \\
& + \log p_\theta(\kdomain{con}=1 \mid g_i,\z_i;\mathcal{B})
  + \sum_{t=1}^{T}
        \log p_\theta(\kdomain{t}=y_{i,t}\mid g_i,\z_i)
\Big].
\end{aligned}
\label{eq:extended-cmim-loss-expanded}
\end{equation}

Algorithm~\ref{algo:cmim} summarizes the full graph-to-SMILES cMIM pretraining loop corresponding to Eq.~\eqref{eq:extended-cmim-loss-expanded}.

\begin{algorithm}[ht]
    \small
    \caption{Graph-to-SMILES cMIM pretraining}
    \label{algo:cmim}
    \begin{algorithmic}[1]
        \REQUIRE Batch $\mathcal{B}=\{\x_i=(g_i,s_i)\}_{i=1}^{B}$; optional self-supervised tasks $\{\operatorname{task}_t\}_{t=1}^{T}$
        \WHILE{not converged}
            \STATE Sample $\z_i \sim q_\theta(\z \mid g_i)$ for all $i \in \{1,\ldots,B\}$
            \STATE Compute SMILES reconstruction terms $\log p_\theta(s_i \mid \z_i)$
            \STATE Compute latent-density terms $\log q_\theta(\z_i \mid g_i)$ and $\log p(\z_i)$
            \STATE Compute contrastive probabilities $p_\theta(\kdomain{con}=1 \mid g_i,\z_i;\mathcal{B})$ using in-batch negatives
            \STATE Compute optional targets $y_{i,t}=\operatorname{task}_t(g_i,s_i)$ and probabilities $p_\theta(\kdomain{t}=y_{i,t}\mid g_i,\z_i)$
            \STATE Form $\hat{\mathcal{L}}_{\text{x-cMIM}}$ using Eq.~\eqref{eq:extended-cmim-loss-expanded}
            \STATE $\theta \leftarrow \theta - \eta \nabla_\theta \hat{\mathcal{L}}_{\text{x-cMIM}}$
        \ENDWHILE
    \end{algorithmic}
\end{algorithm}

\section{Latent-Geometry Evaluation Metrics}\label{app:latent-geometry-metrics}

Section~\ref{sec:results-contrastive-effects} evaluates whether latent nearest neighbors remain close under three external comparison axes: edit distance between canonical SMILES strings~\citep{Weininger1988SMILES}, Morgan-fingerprint distance~\citep{Rogers2010Extended} ($1-$Tanimoto similarity~\citep{Bajusz2015Tanimoto}), and distance between measured ADME-property vectors. Trustworthiness@$k$~\citep{Venna2001Neighborhood} measures whether points close in latent space are also close in the comparison space, so higher values indicate fewer misleading latent neighbors. Precision@$k$ measures the fraction of each molecule's $k$ latent nearest neighbors that also appear among its $k$ nearest neighbors in the comparison space. Violation Rate measures how often latent-near pairs have large comparison-space distance, and conditional $Q_{0.9}$ reports the 90th percentile of comparison-space distance among latent-near pairs. The first two metrics are better when larger; the latter two are better when smaller.

\section{Datasets}\label{app:datasets}

Datasets in this work serve two roles: (i) pretraining the KERMT backbone, where we vary the corpus scale and optionally add in-domain or generated molecules as augmentation, and (ii) downstream tasks including fine-tuning on ADME property prediction (Section~\ref{sec:results-finetune}) and linear probing (Appendix~\ref{app:probing}). The three downstream benchmarks --- Biogen, ExpansionRX, and ChEMBL-MT --- are drawn from experimental ADME assays measured in active drug-discovery campaigns. Realistic ML evaluation for ADME prediction depends on such assay-derived benchmarks because assay noise, protocol variability, and source heterogeneity are difficult to capture with synthetic or public-database-derived targets~\citep{walters2023benchmarks,wognum2024industry}. Table~\ref{tab:pretraining-corpora} shows which datasets are combined in each pretraining configuration.

\paragraph{Pretraining base corpora.}
\begin{itemize}
    \item \textbf{ZINC15\_ChEMBL-11M}: the standard 11M pretraining set regenerated following the protocol of the original GROVER work~\citep{rong2020self}, which combines molecules drawn from ZINC15~\citep{Sterling2015ZINC15} (licensed under custom ZINC/UCSF license) and ChEMBL~\citep{mendez2019chembl} (licensed under CC Attribution-ShareAlike 3.0 Unported license).
    \item \textbf{ZINC15\_ChEMBL-up-208M}: a superset of ZINC15\_ChEMBL-11M that adds approximately 197M additional ZINC15 molecules~\citep{Sterling2015ZINC15}. The additional ZINC15 samples are drawn probabilistically (random seed $42$) from the full ZINC15 catalog ($\sim$1.78B molecules) at a target sample size of 200M, merged with the 11M base, and canonicalized with RDKit (duplicates across sources are removed). This corpus lets us study the effect of pretraining-data volume at a fixed model size.
\end{itemize}

\paragraph{Biogen.}
The Biogen ADME dataset~\citep{fang2023prospective} comprises six endpoints with 3.5K unique molecules. We discarded two of the six endpoints --- human and rat plasma protein binding --- due to their small size (194 and 168 unique molecules respectively, compared to 2{,}172--3{,}086 across the four endpoints we kept). We use the Human Liver Microsomal Clearance (intrinsic), Rat Liver Microsomal Clearance (intrinsic), MDR1-MDCK efflux ratio, and solubility at pH 6.8. We split the dataset into train, validation, and test sets using a Bemis--Murcko scaffold split~\citep{Bemis1996Murcko}, ensuring that molecules with the same scaffold appear in the same split. This split avoids leakage between the splits and tests the ability of our models to generalize to new regions of chemical space unseen during training.

\paragraph{MolMIM-generated augmentation.}
Molecules generated with \texttt{sample\_from\_embeddings.py} from the MolMIM repository~\citep{reidenbach2022improving}, conditioned on embeddings of the Biogen molecules above as seeds, using Gaussian perturbation with noise $\sigma=0.8$ and 50 samples per seed. From the resulting pool of $\sim$103{,}000 unique novel molecules we draw a 30{,}000-molecule subset (random seed $42$) for use as a pretraining augmentation. This augmentation corpus is used only for pretraining, not downstream evaluation.

\paragraph{ExpansionRX (OpenADMET).}
We use the dataset released by ExpansionRX as part of the OpenADMET Blind challenge in January 2026. This dataset comprises 7.6K molecules with nine measured endpoints. These nine endpoints consist of physicochemical properties (LogD and kinetic solubility), clearance properties (human and mouse liver microsomal intrinsic clearance), protein binding (mouse plasma protein binding, mouse brain protein binding, and mouse Gastrocnemius Muscle Binding), and permeability endpoints (Caco-2 efflux ratio and Caco-2 Papp A>B). We use the temporal split provided by the challenge organizers, using molecules synthesized earlier in the drug discovery campaign to train and validate models. Temporal split provides a more realistic estimate of prospective predictive model performance than random or leave-class-out splits~\citep{sheridan2013}. We use molecules synthesized later in the campaign as the test set reflecting retrospective performance in a real-world setting.

\paragraph{ChEMBL-MT.}
The ChEMBL Multi-Task (ChEMBL-MT) dataset is curated from the ChEMBL database~\citep{mendez2019chembl} by Adrian~\etal~\citep{adrian2025multitask}. It consists of 25 endpoints with 114K molecules. Among all endpoints considered in this work, hERG inhibition in ChEMBL-MT is the only toxicity endpoint; the remaining endpoints are ADME assays. We use the Taylor-Butina cluster splits~\citep{Butina1999Clustering} published in Adrian~\etal. In our benchmarks, all endpoints except hERG fall under the ADME category; hERG is included as a toxicity endpoint.

The Biogen and ExpansionRX datasets are single-source datasets with each endpoint measured under very similar experimental conditions. The ChEMBL-MT dataset, however, consists of endpoints aggregated from multiple sources with experiments run in different conditions spanning a large time period. This gives rise to batch effects which cannot be completely accounted for. All three datasets are amenable to multi-task learning. By probing our models on these three datasets, each with its unique characteristics, we attempt to comprehensively and realistically represent the ADME prediction problem. Table~\ref{tab:dataset-overview} summarizes the size, endpoint count, source type, split, and license of the three downstream datasets.

\begin{table}[ht]
    \centering
    \small
    \caption{Summary of the three downstream datasets and splits used to evaluate fine-tuning performance.}
    \begin{tabular}{l|ccc}
        \toprule
                       & Biogen         & ExpansionRX    & ChEMBL-MT                   \\
        \midrule
        Size           & 3.5K molecules & 7.6K molecules & 114K molecules              \\
        \# assays      & 4 endpoints    & 9 endpoints    & 25 endpoints                \\
        Source type    & Single source  & Single source  & Multi source                \\
        Split          & Scaffold split & Temporal split & Taylor-Butina cluster split \\
        License        & MIT License & CC-BY-4.0 & GPL license \\
        \bottomrule
    \end{tabular}
    \label{tab:dataset-overview}
\end{table}


\section{Additional Architecture Figures}\label{app:architectures}

\begin{figure}[ht]
    \centering
    \includegraphics[width=0.65\linewidth]{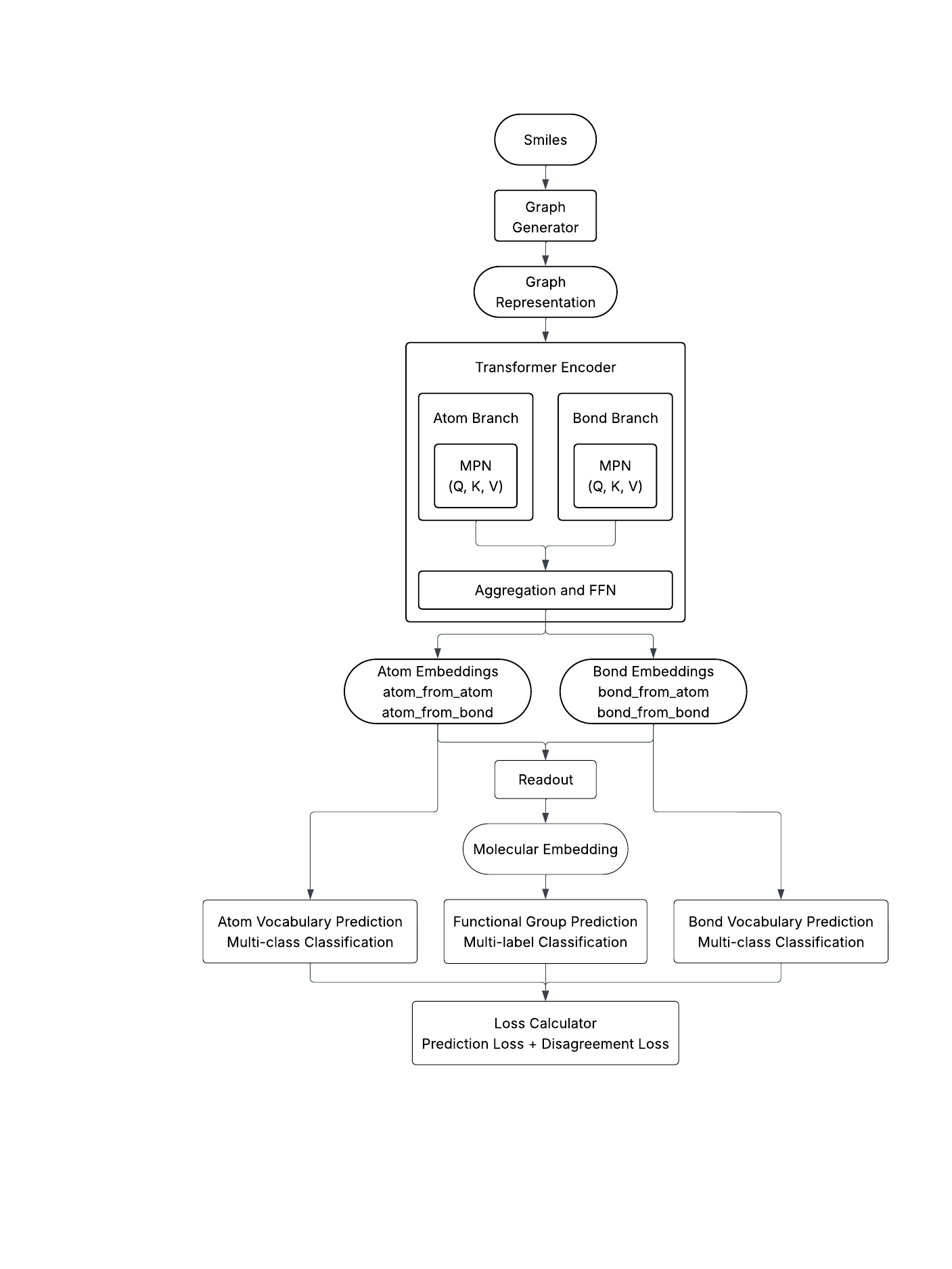}
    \caption{Architecture of the KERMT variant. The encoder (message passing + transformer) is followed directly by the vocabulary-prediction heads and loss calculator; no SMILES decoder or cMIM loss is used.}
    \label{fig:app-arch-kermt}
\end{figure}

\begin{figure}[ht]
    \centering
    \includegraphics[width=0.65\linewidth]{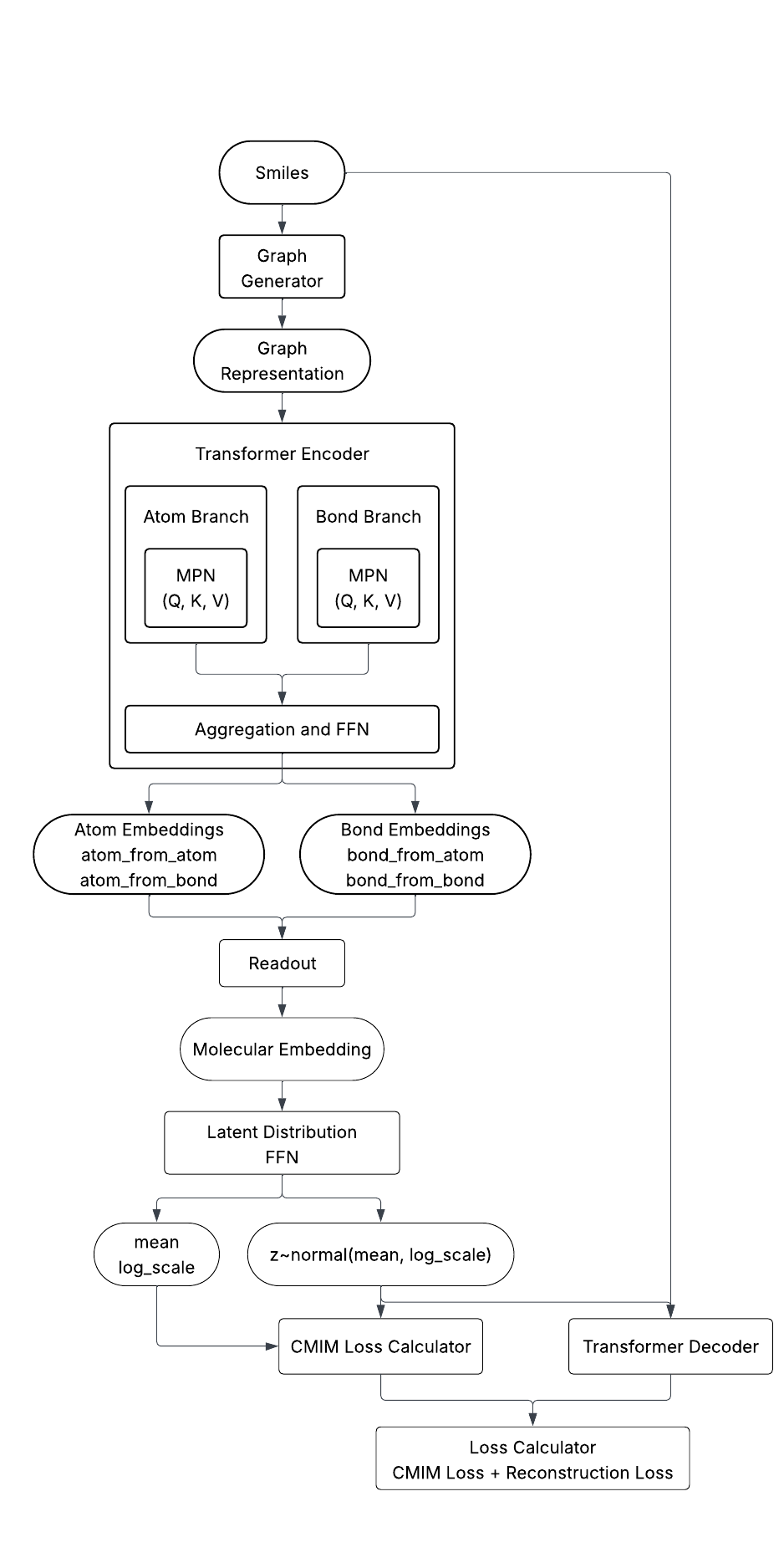}
    \caption{Architecture of cMIM-only. The encoder is followed by a readout, latent-distribution head, and SMILES transformer decoder; the cMIM loss is computed from the latent codes together with the reconstruction loss from the decoder. Vocabulary-prediction heads are not used.}
    \label{fig:app-arch-cmim}
\end{figure}

\begin{figure}[ht]
    \centering
    \includegraphics[width=0.65\linewidth]{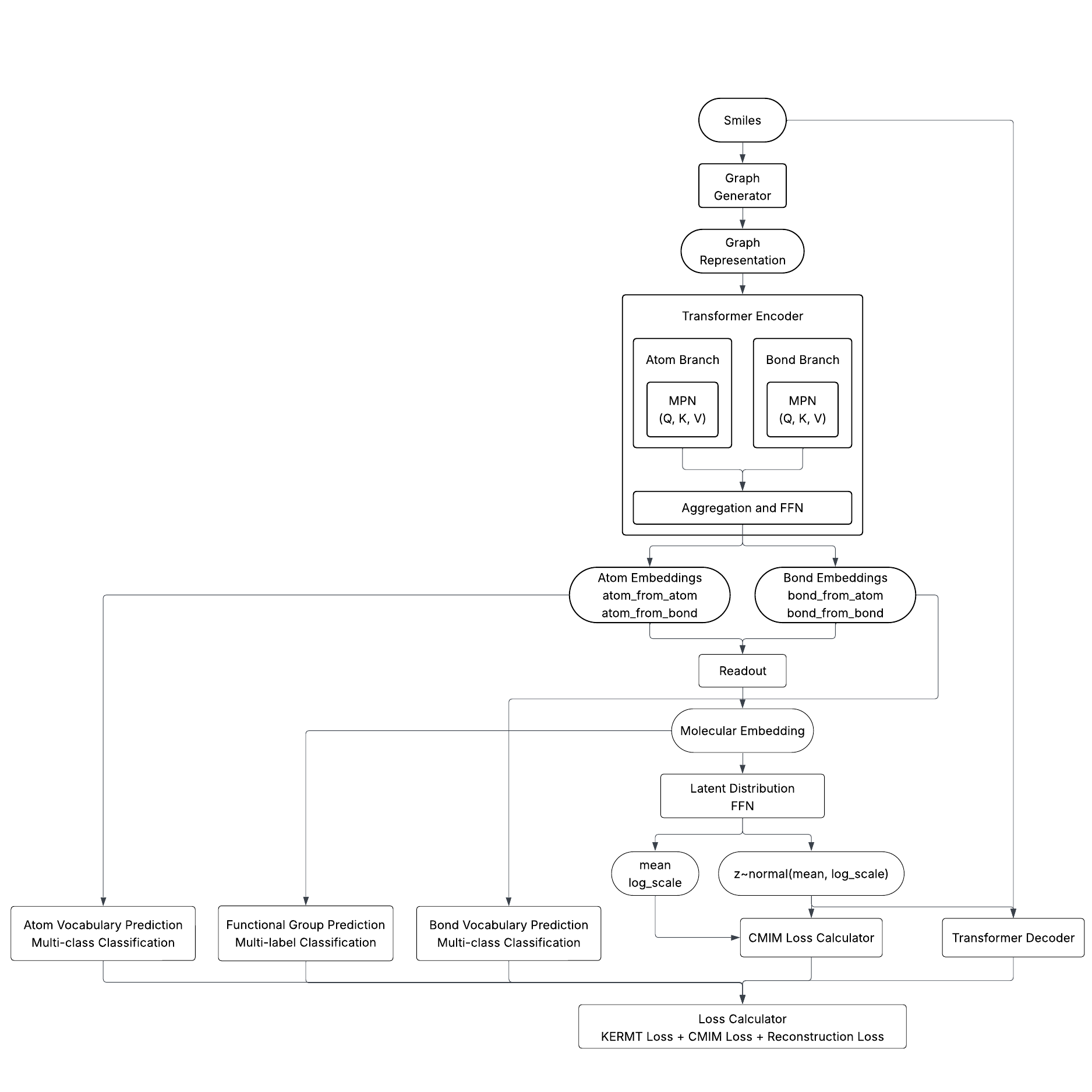}
    \caption{Architecture of Contrastive KERMT, which combines all components of KERMT and cMIM-only: the encoder, the vocabulary-prediction heads, and the SMILES decoder with its cMIM loss are all active in a single pretraining objective.}
    \label{fig:app-arch-hybrid}
\end{figure}

We provide detailed architecture diagrams of the three model variants: KERMT (Figure~\ref{fig:app-arch-kermt}), cMIM-only (Figure~\ref{fig:app-arch-cmim}), and Contrastive KERMT (Figure~\ref{fig:app-arch-hybrid}).

\begin{table}[ht]
    \centering
    \small
    \caption{Comparison of the three model variants. All variants share the same KERMT encoder; cMIM-only and Contrastive KERMT add a SMILES decoder, and Contrastive KERMT additionally retains the vocabulary prediction heads. Parameter counts are reported for the 11M-molecule pretraining configuration; the Contrastive KERMT total varies within $\pm 0.1$M across the augmented configurations in Table~\ref{tab:pretraining-corpora} because the SMILES, atom, and bond vocabularies are rebuilt per pretraining corpus.}
    \label{tab:model-variants}
    \begin{tabular}{lccc}
        \toprule
        Component & KERMT & cMIM-only & Contrastive KERMT \\
        \midrule
        Encoder (message passing + transformer) & \checkmark & \checkmark & \checkmark \\
        SMILES decoder (reconstruction)         &            & \checkmark & \checkmark \\
        Vocabulary prediction heads             & \checkmark &            & \checkmark \\
        \midrule
        Pretraining objective                   & vocab      & cMIM       & vocab + cMIM \\
        Total parameters (pretraining)                        & 56.95M     & 62.04M     & 70.61M \\
        \textbf{Downstream fine-tuning/inference model}  & \multicolumn{3}{c}{\textbf{same KERMT backbone; pretraining-only modules discarded}} \\
        Total parameters (downstream)                        & 56.95M     & 56.95M     & 56.95M \\
        \bottomrule
    \end{tabular}
\end{table}

\section{Implementation Details}\label{app:implementation-details}
The \modelname implementation builds on the KERMT architecture described by Adrian \etal~\citep{adrian2025multitask}.
\makeatletter
\if@preprint
The code is publicly available at \url{https://github.com/NVIDIA-BioNeMo/KERMT}.
\else
For review, we provide an anonymized code package in the supplementary material; the code package will be de-anonymized upon publication.
\fi
\makeatother
\paragraph{Pretraining configuration.}
All pretraining runs use self-supervised targets computed on the fly, and the cMIM and Contrastive KERMT runs additionally use in-batch negatives for the contrastive term; no positive-pair molecular augmentations are required. For variants with a contrastive component, the cMIM temperature is fixed to $\tau=0.1$. Across pretraining variants, runs share the same KERMT encoder backbone and optimizer settings and use random seed $0$; the variants differ in their pretraining objective and pretraining-only modules. Training budgets vary with corpus size: 11M-base configurations, including all augmented 11M variants, are trained for 100 epochs with 20 warmup epochs; the 208M-base run is trained for 6 epochs with 2 warmup epochs; and the 208M-base + Biogen run is trained for 4 epochs with 2 warmup epochs.
For cMIM variants, the posterior head predicts a diagonal Gaussian $q_\theta(\z_i \mid g_i)=\mathcal{N}(\mu_i,\operatorname{diag}(\sigma_i^2))$. Latent samples are drawn with the reparameterization trick. For numerical stability, the variance is clipped below $10^{-6}$. The posterior head is a pretraining-only module and is discarded before downstream fine-tuning.
Loss aggregation follows the probabilistic factors in Eq.~\eqref{eq:extended-cmim-loss}. SMILES reconstruction uses the teacher-forced autoregressive negative log-likelihood with a character-level tokenizer, summed over tokens for each molecule. Gaussian posterior and prior log densities are summed over latent dimensions. KERMT atom, bond, and functional-group objectives are computed as in KERMT, with atom and bond heads averaged before entering the joint objective. Pretraining targets are self-supervised and can be computed on the fly for any molecule. For downstream multi-task fine-tuning, losses are computed only for observed assay labels.

\paragraph{Hardware and compute.}
All pretraining runs use a single node with $8$ NVIDIA A100 GPUs on an internal compute cluster, with synchronous data-parallel training (NCCL backend). Per-GPU batch size is $128$, giving an effective global batch of $1{,}024$ molecules per optimizer step across all variants and corpora. Per-run wall-clock cost is approximately $3$ weeks for 11M-base configurations (100 epochs), $6$ weeks for the 208M-base + Biogen configuration (4 epochs), and $8$ weeks for the 208M-base configuration (6 epochs). Fine-tuning runs are substantially smaller and use a single A100 GPU per (variant, corpus, seed) configuration.
At fixed corpus size, wall-clock pretraining time and GPU-hours were comparable across KERMT, cMIM-only, and Contrastive KERMT. Pretraining-only modules are discarded before fine-tuning, so downstream inference uses the same KERMT encoder/readout interface across pretrained variants.

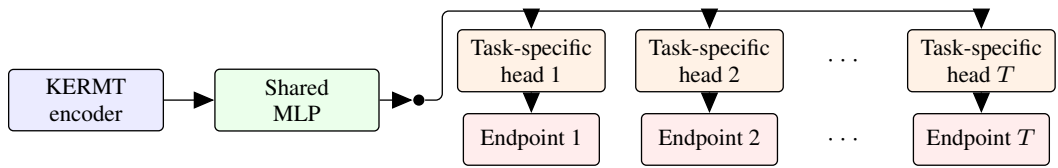
\begin{figure}[ht]
    \centering
    \resizebox{\linewidth}{!}{\begin{tikzpicture}[
    font=\scriptsize,
    x=1cm,
    y=1cm,
    box/.style={
        draw,
        rounded corners=1.5pt,
        align=center,
        inner xsep=4pt,
        inner ysep=3pt,
        minimum height=0.55cm
    },
    core/.style={box, fill=blue!8, minimum width=1.65cm},
    shared/.style={box, fill=green!8, minimum width=1.75cm},
    head/.style={box, fill=orange!10, minimum width=1.55cm},
    endpoint/.style={box, fill=red!7, minimum width=1.45cm},
    junction/.style={circle, fill=black, inner sep=1.2pt},
    arrow/.style={->, line width=0.4pt}
]
    \node[core] (enc) at (0.0,0.0) {KERMT\\encoder};
    \node[shared] (shared) at (2.25,0.0) {Shared\\MLP};
    \node[junction] (split) at (3.55,0.0) {};
    \coordinate (busStart) at (3.95,0.95);
    \coordinate (busEnd) at (9.55,0.95);
    \coordinate (tap1) at (4.75,0.95);
    \coordinate (tap2) at (6.65,0.95);
    \coordinate (tapT) at (9.55,0.95);

    \node[head] (h1) at (4.75,0.42) {Task-specific\\head 1};
    \node[head] (h2) at (6.65,0.42) {Task-specific\\head 2};
    \node at (8.10,0.42) {$\cdots$};
    \node[head] (hT) at (9.55,0.42) {Task-specific\\head $T$};

    \node[endpoint] (e1) at (4.75,-0.42) {Endpoint 1};
    \node[endpoint] (e2) at (6.65,-0.42) {Endpoint 2};
    \node at (8.10,-0.42) {$\cdots$};
    \node[endpoint] (eT) at (9.55,-0.42) {Endpoint $T$};

    \draw[arrow] (enc) -- (shared);
    \draw[arrow] (shared) -- (split);
    \draw[line width=0.4pt, rounded corners=2pt] (split) -- ++(0.25,0) |- (busStart) -- (busEnd);
    \draw[arrow] (tap1) -- (h1.north);
    \draw[arrow] (tap2) -- (h2.north);
    \draw[arrow] (tapT) -- (hT.north);
    \draw[arrow] (h1) -- (e1);
    \draw[arrow] (h2) -- (e2);
    \draw[arrow] (hT) -- (eT);
\end{tikzpicture}}
    \caption{Task-specific MLP heads, one per endpoint, on top of a shared MLP trunk for downstream fine-tuning.}
    \label{fig:task_specific_heads_schematic}
\end{figure}

\paragraph{Fine-tuning protocols.}
The task-specific fine-tuning configuration uses a 2-layer shared MLP trunk followed by a 3-layer task-specific MLP head per endpoint (Figure~\ref{fig:task_specific_heads_schematic}). The default configuration follows the standard KERMT~\citep{rong2020self} fine-tuning recipe with a single 3-layer shared MLP and no per-endpoint specialization. All fine-tuning MLPs use hidden size 700. For both configurations, each pretrained backbone initializes property prediction on the three ADME-focused benchmarks of Appendix~\ref{app:datasets}, and we fine-tune end-to-end for 100 epochs. Pretraining-only modules are discarded before fine-tuning: downstream models use the same KERMT encoder/readout interface regardless of whether pretraining included cMIM. Scores are averaged over five seeds $\{0, 1, 2, 3, 4\}$, except for ChEMBL-MT where we use four runs comprising two cluster-split folds and two seeds $\{0, 1\}$ per fold. We ran the default protocol for the KERMT baseline on all three benchmarks as a no-task-specific reference; for cMIM-only and Contrastive KERMT, we ran the default protocol only on Biogen because task-specific layers consistently improved Biogen performance and were therefore adopted for ExpansionRX and ChEMBL-MT.

\paragraph{Baseline implementations.}
Fixed-descriptor baselines include Morgan fingerprints computed with RDKit~\citep{rdkit_2025_09_4} (licensed under BSD 3-Clause License) at radius 2, folded into 1024 bits and passed to a 3-layer feed-forward network, and 200 RDKit 2D descriptors normalized with \texttt{descriptastorus}~\citep{descriptastorus} (licensed under BSD 3-Clause License). Chemprop uses the directed message passing neural network implementation in the Chemprop package~\citep{graff2025} (licensed under MIT License), with external-dataset hyperparameters from Adrian \etal~\citep{adrian2024}. MolCLR is a graph isomorphism network trained contrastively through augmentations; we adapt the official MolCLR codebase~\citep{Wang2022MolCLRRepo} (licensed under MIT License) for multi-task prediction. KPGT, a line graph transformer that incorporates RDKit descriptor and fingerprint knowledge, uses the official codebase~\citep{Li2023KPGTRepo} (licensed under Apache License 2.0) with default hyperparameters. The KERMT baseline shares the \modelname backbone architecture and uses hyperparameters from Adrian \etal~\citep{adrian2025multitask}.

\begin{figure}[ht]
    \centering
    \includegraphics[width=0.9\linewidth]{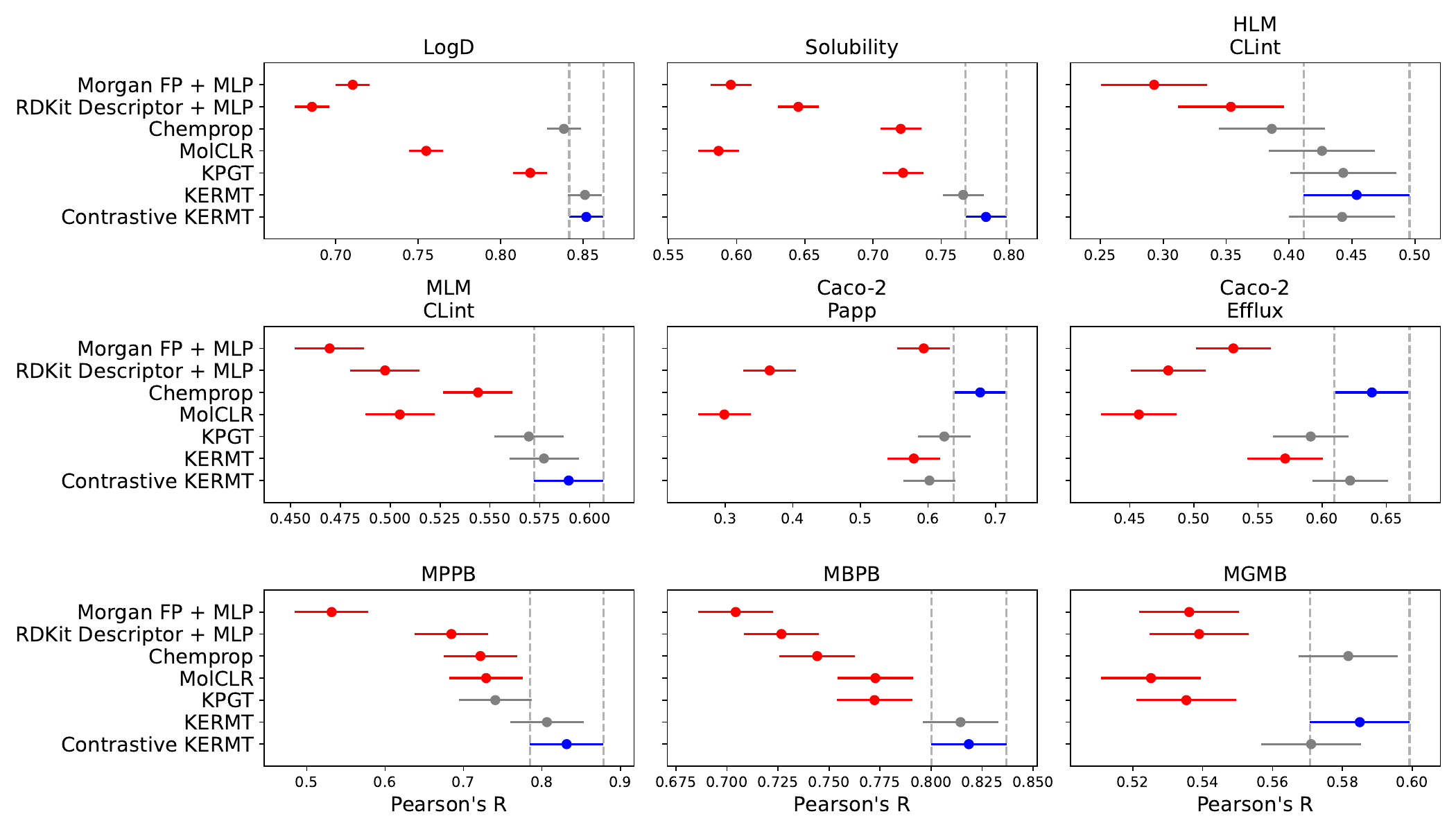}
    \caption{Tukey HSD plot comparing \modelname model performance on ExpansionRX dataset with all baselines. The ANOVA p-values were $<10^{-4}$ for all assays. Blue: best method, Gray: statistically indistinguishable from the best method, Red: statistically worse than the best method.}
    \label{fig:comp_with_baselines-exprx}
\end{figure}

\begin{figure}[ht]
    \centering
    \includegraphics[width=0.6\linewidth]{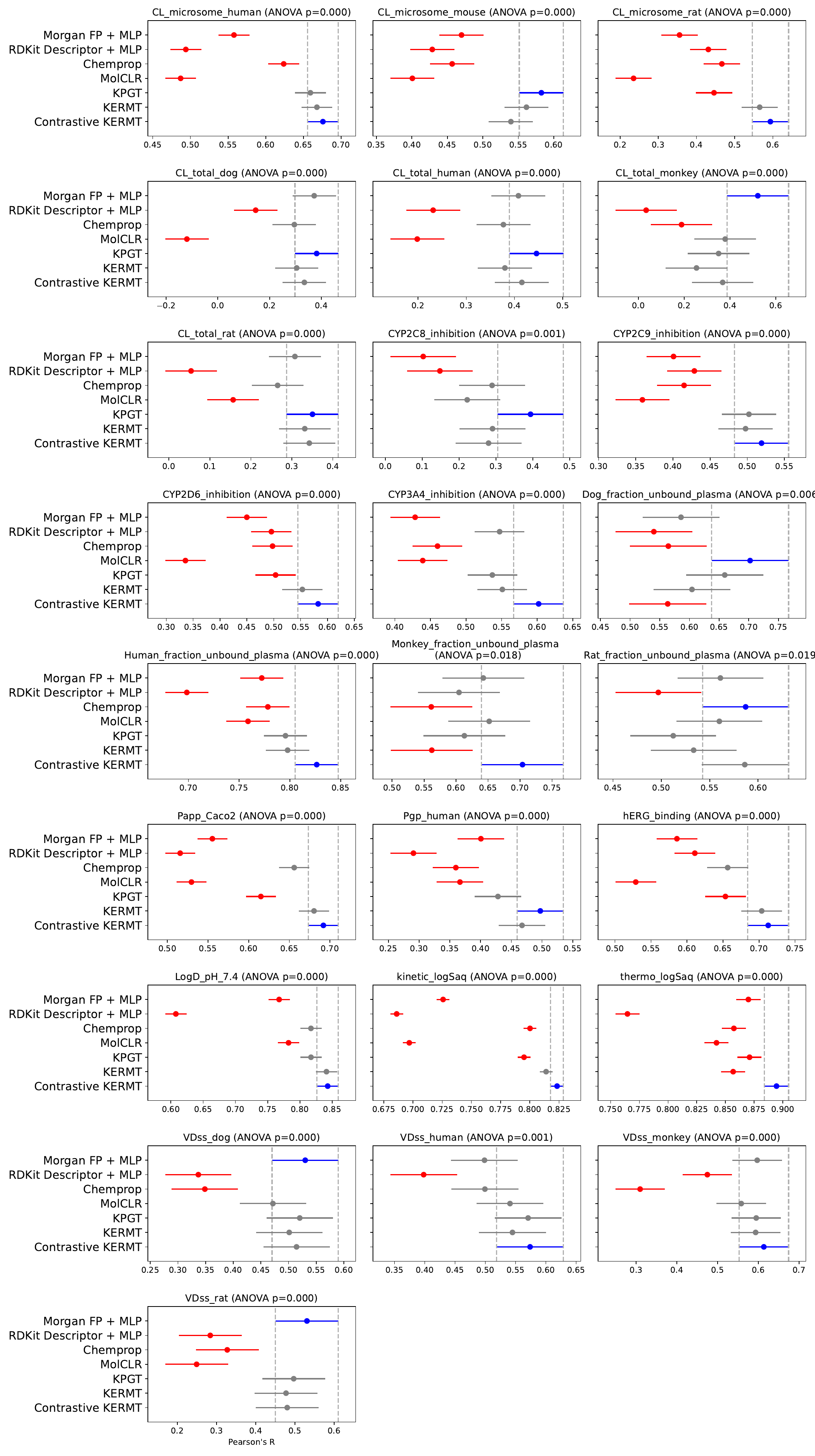}
    \caption{Tukey HSD plot comparing \modelname model performance on ChEMBL-MT dataset with all baselines. Blue: best method, Gray: statistically indistinguishable from the best method, Red: statistically worse than the best method.}
    \label{fig:comp_with_baselines-chembl-mt}
\end{figure}

\paragraph{Statistical testing.}
For cross-baseline comparisons, we test differences across models separately for each endpoint using one-way Analysis of variance (ANOVA) over independent training runs, followed by Tukey's Honest Significant Difference (HSD) test from the \texttt{statsmodels} Python package~\citep{seabold2010statsmodels}. The run-level samples correspond to random seeds for Biogen and ExpansionRX and seed/fold combinations for ChEMBL-MT. Tukey HSD accounts for multiple pairwise model comparisons~\citep{Ash2025practically}, and plotted confidence intervals are halfwidths computed using Tukey's $Q$ critical value. We report p-values in scientific notation or as thresholds when appropriate. These cross-baseline ANOVA/Tukey tests are distinct from the two-sample $t$-tests used for the significant-endpoint conditional averages in Appendix~\ref{app:finetune-significance}. The resulting Tukey HSD comparison plots are shown in Figure~\ref{fig:comp_with_baselines-biogen} (Biogen, in Section~\ref{sec:comp_with_baselines}), Figure~\ref{fig:comp_with_baselines-exprx} (ExpansionRX), and Figure~\ref{fig:comp_with_baselines-chembl-mt} (ChEMBL-MT).

\section{Pretraining Data Source Distribution Analysis}\label{app:embedding-viz}

To understand how consistent the data sources used across pretraining and downstream evaluation (Section~\ref{sec:experiments} and Section~\ref{sec:downstream}) are in chemical space, we run a nearest-neighbor source-composition analysis on molecule embeddings. For each molecule we retrieve its top-20 cosine nearest neighbors (excluding self) and count the proportion that belongs to each data source. Averaging these per-molecule proportions over molecules grouped by source yields a confusion matrix whose $i$-th row gives the average neighbor-source composition for molecules of source $i$: the diagonal measures self-source purity (a high value indicates a well-separated source) and off-diagonal entries measure how strongly two sources share chemical space.

\begin{figure}[t]
    \centering
    \includegraphics[width=0.75\linewidth]{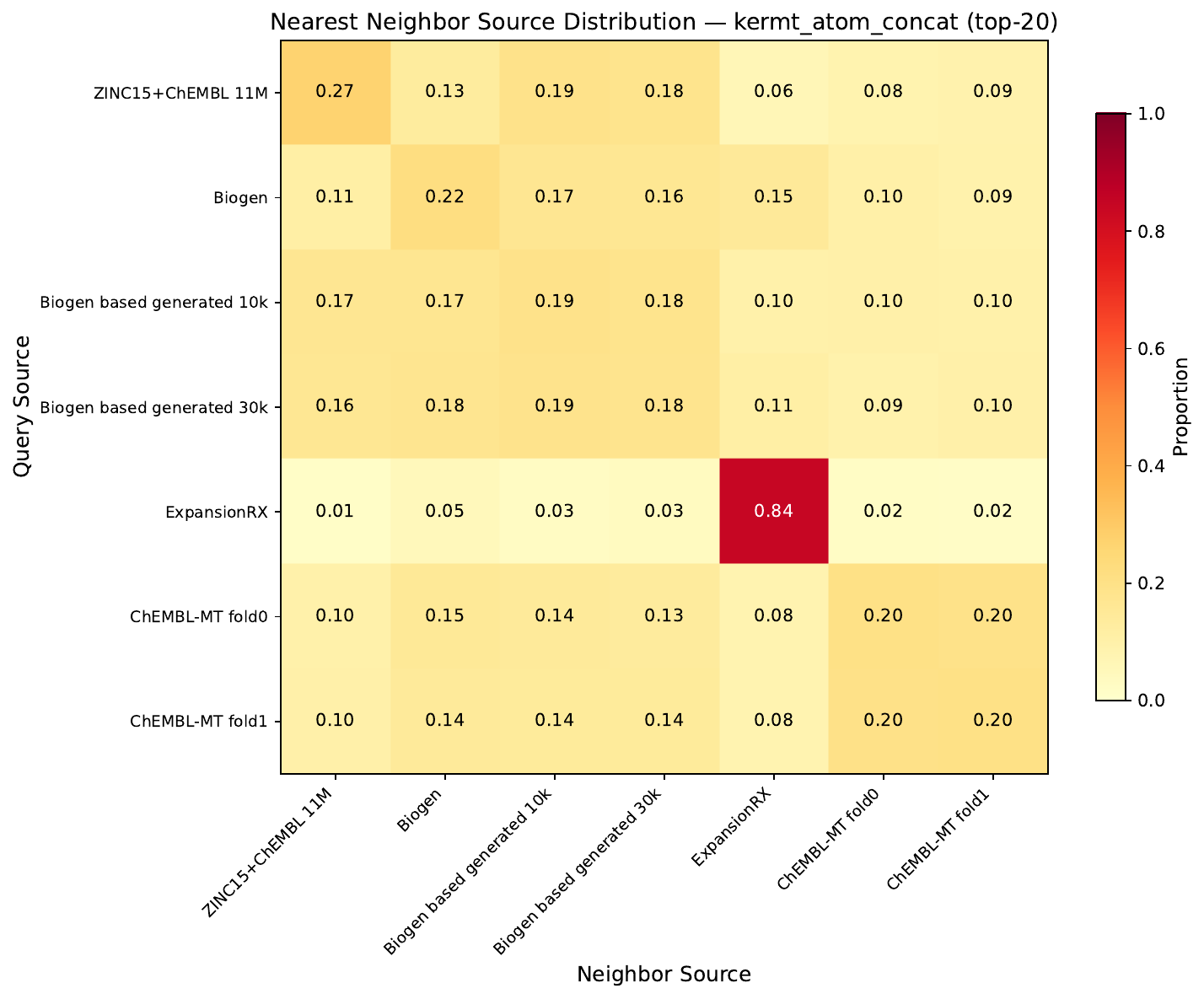}
    \caption{Top-20 nearest-neighbor source composition in Contrastive KERMT embedding space. Each row gives the average source composition of the top-20 cosine neighbors of molecules from that source; the diagonal is self-source purity. ExpansionRX separates from every other source while the remaining sources mix with each other.}
    \label{fig:kermt-confusion}
\end{figure}

\begin{figure}[t]
    \centering
    \includegraphics[width=0.75\linewidth]{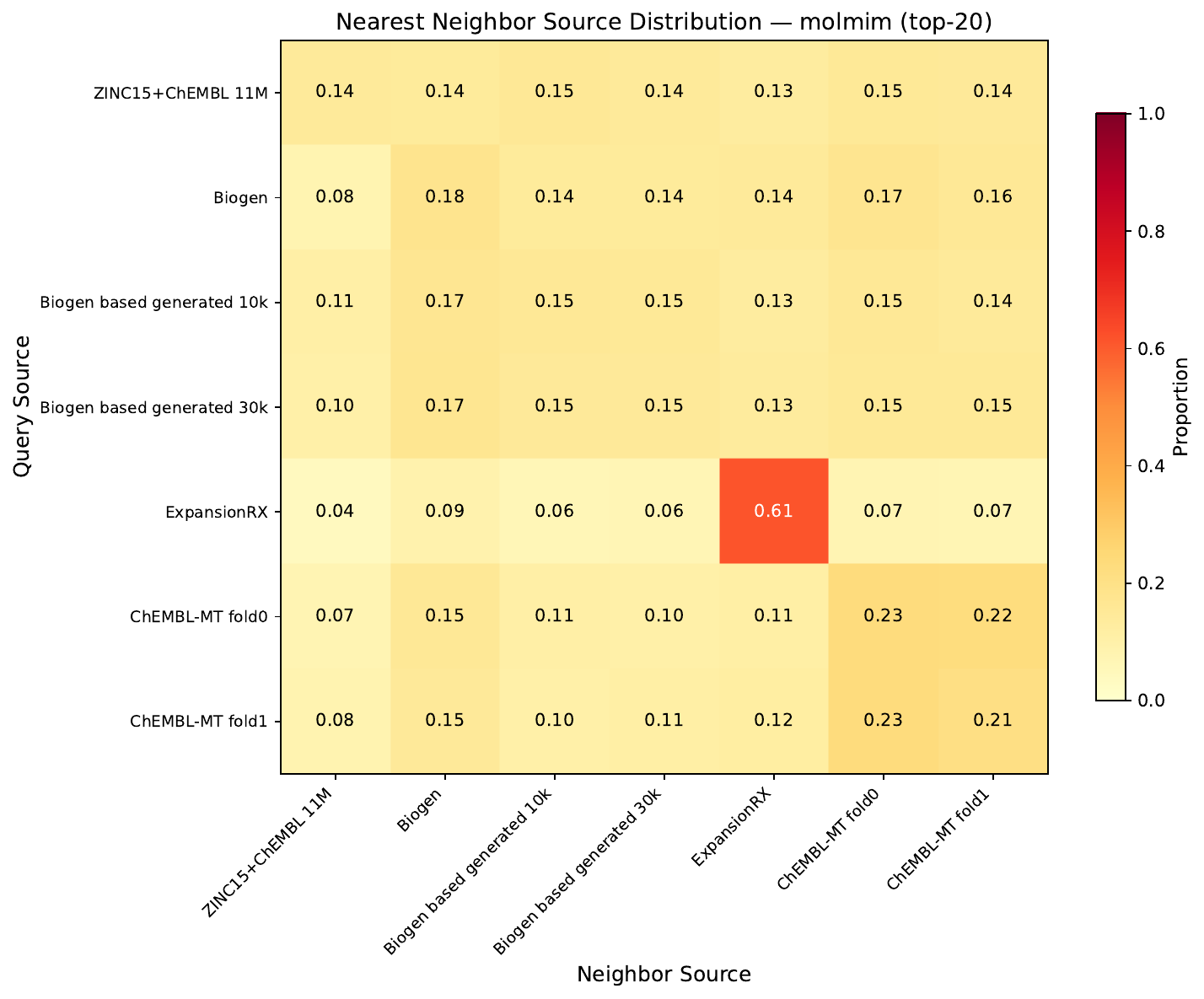}
    \caption{Top-20 nearest-neighbor source composition in MolMIM~\citep{reidenbach2022improving} embedding space, included as an independent reference. The same qualitative patterns hold: MolMIM-generated pools overlap with their Biogen seeds, and ExpansionRX stands apart from the other sources.}
    \label{fig:molmim-confusion}
\end{figure}

We report the analysis in two embedding spaces. Figure~\ref{fig:kermt-confusion} uses Contrastive KERMT embeddings and is directly relevant to our pretraining setup. Figure~\ref{fig:molmim-confusion} uses MolMIM~\citep{reidenbach2022improving} embeddings, included as an independent reference latent space learned from a different objective on a different corpus.

Two patterns are worth highlighting, both visible in both embedding spaces:
\begin{itemize}
    \item \textbf{MolMIM-generated molecules cluster with their Biogen seeds.} The neighbors of each generated subset are dominated by the Biogen seed set and by the other generated subsets, confirming that our MolMIM-based augmentation (Section~\ref{sec:experiments}) does not drift far from the seed distribution.
    \item \textbf{ExpansionRX behaves as an outlier relative to the rest of the corpus.} Its self-source proportion is unusually high, and molecules from other sources rarely retrieve ExpansionRX neighbors. The remaining sources (Biogen, its MolMIM-generated variants, and ChEMBL-MT) mix with each other comparatively evenly, while ExpansionRX occupies a distinct region of chemical space---effectively an out-of-distribution slice relative to the other ADME-adjacent corpora. This pattern is robust across both embedding spaces, indicating that it is a property of the data rather than of any particular encoder.
\end{itemize}

\section{Per-endpoint Fine-tuning Results}\label{app:finetune}

\begin{figure}[t]
    \centering
    \includegraphics[width=\linewidth]{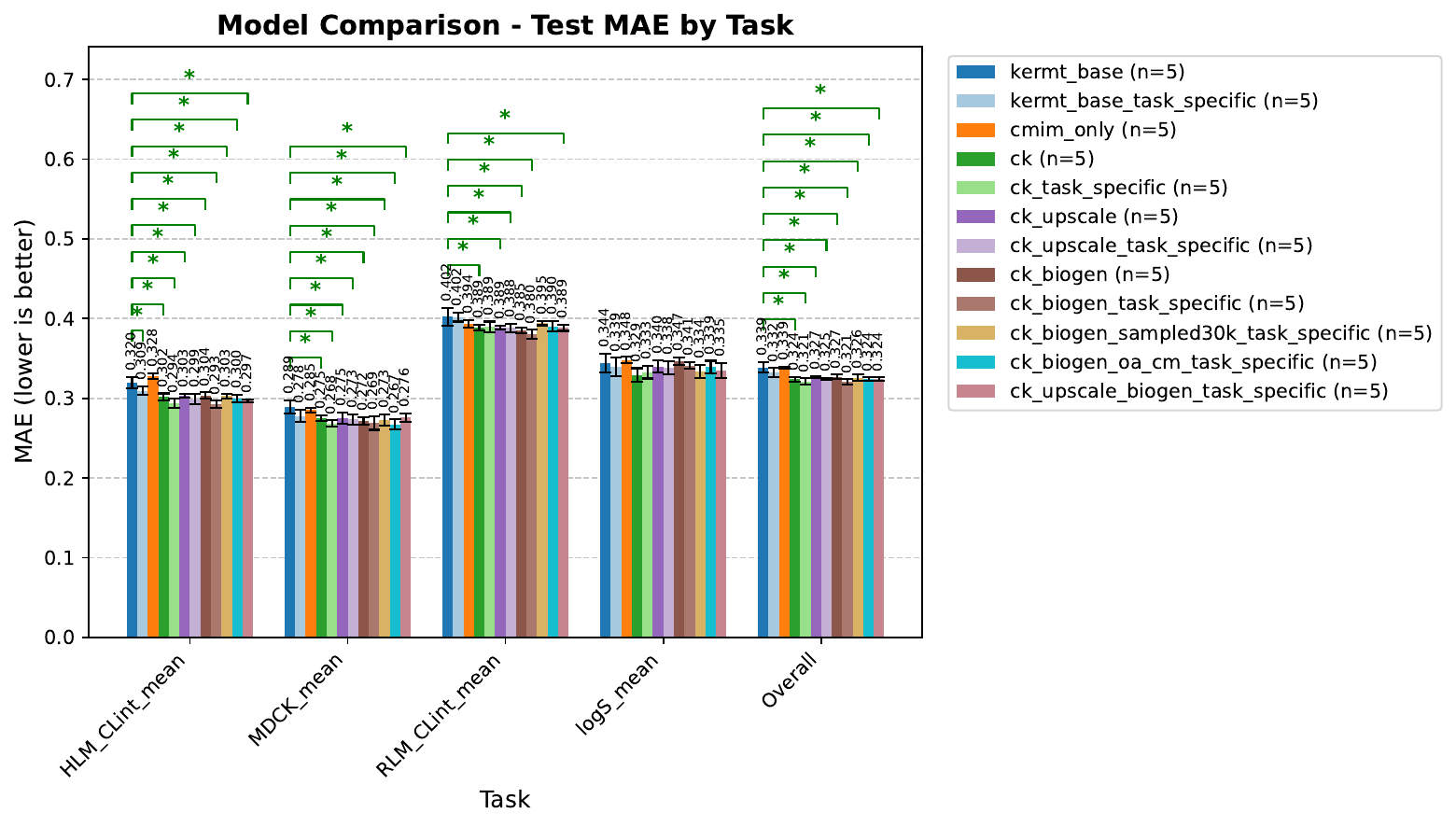}
    \caption{Fine-tuning performance on the Biogen ADME dataset across model variants and pretraining corpora. The KERMT baseline (\texttt{kermt\_base}) is shown first in each panel for reference.}
    \label{fig:finetune-biogen}
\end{figure}

\begin{figure}[t]
    \centering
    \includegraphics[width=\linewidth]{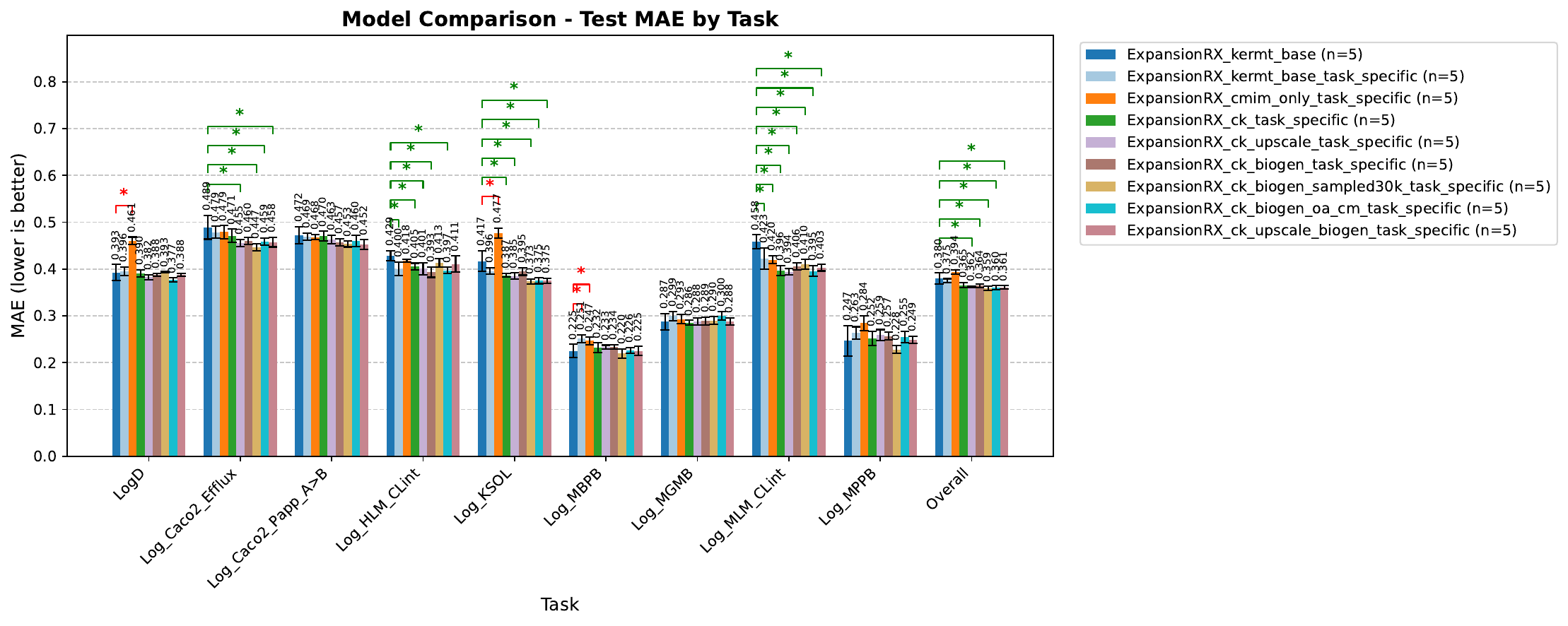}
    \caption{Fine-tuning performance on ExpansionRX across model variants and pretraining corpora. Trends are consistent with the Biogen results: the Contrastive KERMT configurations outperform the single-objective backbones, and adding ADME-adjacent corpora to the pretraining mix provides further gains.}
    \label{fig:finetune-openadmet}
\end{figure}

\begin{figure}[p]
    \centering
    \includegraphics[angle=90,height=0.92\textheight]{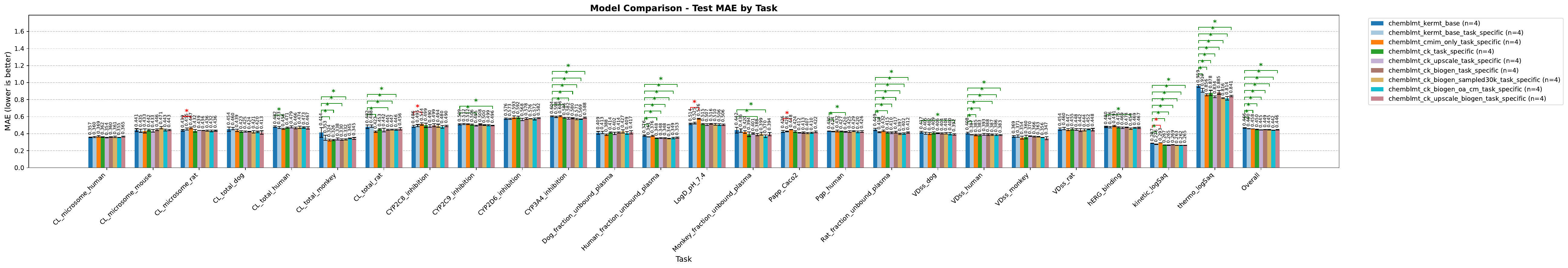}
    \caption{Fine-tuning performance on ChEMBL-MT across model variants and pretraining corpora.}
    \label{fig:finetune-chemblmt}
\end{figure}

Table~\ref{tab:finetune-improvement} reports the raw mean MAE across endpoints for every (pretraining variant, corpus, fine-tuning protocol) configuration we evaluated. Figures~\ref{fig:finetune-biogen}, \ref{fig:finetune-openadmet}, and~\ref{fig:finetune-chemblmt} report the per-endpoint fine-tuning performance on Biogen, ExpansionRX, and ChEMBL-MT respectively. The ranking across variants is consistent across all three benchmarks despite the differing task counts.

\begin{table}[ht]
    \centering
    \small
    \caption{Mean MAE across endpoints for each (pretraining variant, fine-tuning protocol) configuration. Lower is better; bold marks the best (lowest) MAE per benchmark column. Compact corpus names: ``11M base'' = ZINC15\_ChEMBL-11M, ``208M base'' = ZINC15\_ChEMBL-up-208M; OA = ExpansionRX, CM = ChEMBL-MT (full corpus definitions in Table~\ref{tab:pretraining-corpora}). Protocol, seed, and missing-run details are provided in Appendix~\ref{app:implementation-details}.}
    \label{tab:finetune-improvement}
    \resizebox{\linewidth}{!}{%
    \begin{tabular}{lllccc}
        \toprule
        Pretraining objective         & Pretraining corpus              & Fine-tuning     & Biogen          & ExpansionRX       & ChEMBL-MT       \\
        \midrule
        vocab (\texttt{kermt\_base})  & 11M base                        & default         & 0.339           & 0.380           & 0.466           \\
        vocab (\texttt{kermt\_base})  & 11M base                        & task-specific   & 0.332              & 0.375           & 0.460           \\
        \midrule
        cMIM-only                     & 11M base                        & default         & 0.339           & NA              & NA              \\
        cMIM-only                     & 11M base                        & task-specific   & NA              & 0.394           & 0.458           \\
        \midrule
        Contrastive KERMT             & 11M base                        & default         & 0.324           & NA              & NA              \\
        Contrastive KERMT             & 11M base                        & task-specific   & \textbf{0.321}  & 0.365           & 0.450           \\
        Contrastive KERMT             & 208M base                       & default         & 0.327           & NA              & NA              \\
        Contrastive KERMT             & 208M base                       & task-specific   & 0.325           & 0.362           & 0.445           \\
        Contrastive KERMT             & 11M base + Biogen               & default         & 0.327           & NA              & NA              \\
        Contrastive KERMT             & 11M base + Biogen               & task-specific   & \textbf{0.321}  & 0.364           & 0.449           \\
        Contrastive KERMT             & 208M base + Biogen              & task-specific   & 0.324           & 0.361           & 0.446           \\
        Contrastive KERMT             & 11M base + Biogen + 30k MolMIM  & task-specific   & 0.326           & \textbf{0.359}  & 0.445           \\
        Contrastive KERMT             & 11M base + Biogen + OA + CM     & task-specific   & 0.324           & 0.360           & \textbf{0.442}  \\
        \bottomrule
    \end{tabular}%
    }
\end{table}

\subsection{Significant endpoints contributing to aggregate improvements}\label{app:finetune-significance}

The aggregate improvements reported in the abstract and Section~\ref{sec:results-finetune} (e.g., $+7.6\%$ / $+9.9\%$ / $+9.5\%$ on Biogen / ExpansionRX / ChEMBL-MT) are significant-endpoint conditional averages: they average per-endpoint percent change only over endpoints with a statistically significant difference (two-sample $t$-test, $p < 0.05$) versus the corresponding reported \texttt{kermt\_base} baseline. Table~\ref{tab:finetune-improvement} reports the corresponding all-endpoint macro mean MAE for every evaluated configuration. Tables~\ref{tab:per-endpoint-biogen}, \ref{tab:per-endpoint-exprx}, and \ref{tab:per-endpoint-chemblmt} list the significant endpoints and their per-endpoint $\Delta\%$ for the three abstract-headline comparisons.

\begin{table}[ht]
    \centering
    \small
    \caption{Biogen: significant endpoints contributing to the $+7.6\%$ aggregate for \texttt{ck\_task\_specific} vs.\ \texttt{kermt\_base} (default-protocol baseline).}
    \label{tab:per-endpoint-biogen}
    \begin{tabular}{lc}
        \toprule
        Endpoint    & $\Delta\%$ \\
        \midrule
        HLM\_CLint  & $+8.12\%$ \\
        MDCK        & $+7.15\%$ \\
        \midrule
        \textbf{Aggregate (mean)} & $\mathbf{+7.63\%}$ \\
        \bottomrule
    \end{tabular}
\end{table}

\begin{table}[ht]
    \centering
    \small
    \caption{ExpansionRX: significant endpoints contributing to the $+9.9\%$ aggregate for \texttt{ck\_biogen\_sampled30k\_task\_specific} vs.\ \texttt{kermt\_base} (default-protocol baseline).}
    \label{tab:per-endpoint-exprx}
    \begin{tabular}{lc}
        \toprule
        Endpoint                & $\Delta\%$ \\
        \midrule
        Log\_MLM\_CLint         & $+10.5\%$ \\
        Log\_KSOL               & $+10.4\%$ \\
        Log\_Caco2\_Efflux      & $+8.7\%$  \\
        \midrule
        \textbf{Aggregate (mean)} & $\mathbf{+9.87\%}$ \\
        \bottomrule
    \end{tabular}
\end{table}

\begin{table}[ht]
    \centering
    \small
    \caption{ChEMBL-MT: significant endpoints contributing to the $+9.5\%$ aggregate for \texttt{ck\_biogen\_oa\_cm\_task\_specific} vs.\ \texttt{kermt\_base} (default-protocol baseline).}
    \label{tab:per-endpoint-chemblmt}
    \begin{tabular}{lc}
        \toprule
        Endpoint                          & $\Delta\%$ \\
        \midrule
        Monkey\_fraction\_unbound\_plasma & $+16.4\%$ \\
        thermo\_logSaq                    & $+15.1\%$ \\
        Rat\_fraction\_unbound\_plasma    & $+10.5\%$ \\
        kinetic\_logSaq                   & $+8.5\%$  \\
        Human\_fraction\_unbound\_plasma  & $+7.4\%$  \\
        CL\_total\_rat                    & $+7.3\%$  \\
        CYP3A4\_inhibition                & $+5.7\%$  \\
        VDss\_human                       & $+5.3\%$  \\
        \midrule
        \textbf{Aggregate (mean)} & $\mathbf{+9.51\%}$ \\
        \bottomrule
    \end{tabular}
\end{table}

\section{Linear Probing on Frozen Embeddings}\label{app:probing}

Beyond the contrastive-geometry analysis (Section~\ref{sec:results-contrastive-effects}) and the cross-baseline and ablation evaluations (Sections~\ref{sec:comp_with_baselines} and~\ref{sec:results-finetune}), we also assess the quality of the pretrained representations via a linear-probing analysis on frozen molecule-level embeddings. Probing isolates how much chemistry signal is already linearly accessible from the pretrained latent space, independent of any downstream-prediction head, and lets us check whether the same configurations that win on fine-tune also win on this representation-quality measurement.

\paragraph{Probe protocol.}
We extract frozen molecule-level embeddings from the Biogen and ExpansionRX datasets (Appendix~\ref{app:datasets}) and fit simple linear models on 17 basic molecular descriptors computed with RDKit (Table~\ref{tab:probing-tasks}): 9 classification tasks (logistic regression, accuracy) and 8 regression tasks (ridge regression, $R^2$). Each probe is fit with 5-fold cross-validation and we report the cross-validation mean per task. Probing evaluates all four encoder readouts independently (mean-pooled to molecule level), then aggregates over the four readouts as described in the caption of Table~\ref{tab:probing-summary}.

\begin{table}[ht]
    \centering
    \small
    \caption{Linear probing tasks derived from RDKit descriptors. Classification targets are two binary druglikeness flags (Lipinski, Veber) and seven integer-count descriptors used directly as discrete-valued multiclass labels; regression targets are continuous descriptors.}
    \label{tab:probing-tasks}
    \begin{tabular}{ll}
        \toprule
        Classification (9)   & Regression (8) \\
        \midrule
        NumHDonors           & MolecularWeight \\
        NumHAcceptors        & LogP \\
        NumRotatableBonds    & TPSA \\
        NumRings             & NumAtoms \\
        NumAromaticRings     & NumHeavyAtoms \\
        NumHeteroatoms       & NumBonds \\
        NumStereocenters     & FractionCSP3 \\
        Lipinski             & BertzCT \\
        Veber                & \\
        \bottomrule
    \end{tabular}
\end{table}

\begin{table}[ht]
    \centering
    \small
    \caption{Linear probing results across the 17 descriptor tasks of Table~\ref{tab:probing-tasks}, evaluated on Biogen-source and ExpansionRX-source molecule embeddings. \textit{Wins} counts per-dataset best-in-class outcomes across all (model, readout) combinations. \textit{Avg} columns average over the four readouts; \textit{Best} columns report the score of the single best-performing readout. Bold marks the best entry per column within each dataset; ties are bolded jointly.}
    \label{tab:probing-summary}
    \begin{tabular}{lccccccc}
        \toprule
        & & \multicolumn{3}{c}{Avg over readouts} & \multicolumn{3}{c}{Best readout} \\
        \cmidrule(lr){3-5} \cmidrule(lr){6-8}
        Model & Wins & Clf & Reg & All & Clf & Reg & All \\
        \midrule
        \multicolumn{8}{l}{\textit{Biogen-source embeddings (17 tasks):}} \\
        \texttt{kermt\_base}                           & 1.0           & 0.832             & 0.755             & 0.796             & 0.868             & 0.867          & 0.868 \\
        \texttt{cmim\_only}                                  & 0.0           & 0.706             & 0.703             & 0.704             & 0.842             & 0.906          & 0.872 \\
        \texttt{ck}                                & 0.0           & 0.855             & 0.881             & 0.867             & 0.882             & 0.932          & 0.905 \\
        \texttt{ck\_upscale}                       & 0.0           & 0.851             & 0.883             & 0.866             & 0.879             & 0.937          & 0.906 \\
        \texttt{ck\_biogen}                        & 1.0           & \textbf{0.857}    & 0.883             & 0.870             & \textbf{0.885}    & 0.937          & 0.909 \\
        \texttt{ck\_upscale\_biogen}               & \textbf{9.0}  & 0.850             & \textbf{0.895}    & \textbf{0.871}    & 0.884             & \textbf{0.949} & \textbf{0.915} \\
        \texttt{ck\_biogen\_sampled30k}            & 3.0           & 0.856             & 0.882             & 0.868             & 0.882             & 0.946          & 0.912 \\
        \texttt{ck\_biogen\_oa\_cm}                & 3.0           & \textbf{0.857}    & 0.846             & 0.852             & 0.882             & 0.946          & 0.912 \\
        \midrule
        \multicolumn{8}{l}{\textit{ExpansionRX-source embeddings (17 tasks):}} \\
        \texttt{kermt\_base}                           & 1.0           & 0.876             & 0.649             & 0.756             & 0.897             & 0.781          & 0.835 \\
        \texttt{cmim\_only}                                  & 0.0           & 0.769             & 0.542             & 0.649             & 0.872             & 0.804          & 0.836 \\
        \texttt{ck}                                & 0.0           & \textbf{0.888}    & 0.735             & 0.807             & 0.903             & 0.827          & 0.863 \\
        \texttt{ck\_upscale}                       & 0.0           & 0.884             & 0.745             & 0.810             & 0.900             & 0.850          & 0.873 \\
        \texttt{ck\_biogen}                        & 3.0           & 0.887             & \textbf{0.775}    & \textbf{0.828}    & 0.904             & 0.854          & 0.878 \\
        \texttt{ck\_upscale\_biogen}               & 1.0           & 0.886             & 0.744             & 0.811             & 0.907             & 0.834          & 0.868 \\
        \texttt{ck\_biogen\_sampled30k}            & 4.0           & 0.887             & 0.763             & 0.821             & \textbf{0.908}    & 0.838          & 0.871 \\
        \texttt{ck\_biogen\_oa\_cm}                & \textbf{8.0}  & 0.886             & 0.738             & 0.808             & 0.904             & \textbf{0.875} & \textbf{0.889} \\
        \bottomrule
    \end{tabular}
\end{table}

\paragraph{Contrastive KERMT dominates the linearly-accessible signal.}
On both embedding sources Contrastive KERMT (\texttt{ck}) lifts average performance well above KERMT (\texttt{kermt\_base}): from $0.796 \to 0.867$ on Biogen-source and $0.756 \to 0.807$ on ExpansionRX-source embeddings. cMIM-only collapses on probing, confirming the same pattern observed in fine-tuning (Section~\ref{sec:results-finetune}): the cMIM contrastive signal is informative only when combined with the vocabulary-prediction objectives, not as a standalone replacement.

\paragraph{The probing-best configuration is dataset-specific.}
On Biogen-source embeddings, the upscale-plus-Biogen configuration (\texttt{ck\_upscale\_biogen}) wins the most tasks ($9/17$, the highest count anywhere in the table) and produces the best regression readout ($R^2 = 0.949$) and the best readout-averaged regression score ($0.895$). On ExpansionRX-source embeddings, the pooled configuration (\texttt{ck\_biogen\_oa\_cm}) wins the most tasks ($8/17$) and produces the best regression readout ($R^2 = 0.875$). The two corpus designs map onto the two evaluation sets, though with different sharpness: on Biogen-source probing, \texttt{ck\_upscale\_biogen} (208M base + Biogen) is the dominant winner across the win count and most aggregate columns; on the more chemically-diverse ExpansionRX source the picture is mixed, with the pooled \texttt{ck\_biogen\_oa\_cm} leading the win-count and best-readout columns and the Biogen-only \texttt{ck\_biogen} leading the readout-averaged columns.

\paragraph{Probing supports the fine-tune model selection.}
The pooled \texttt{ck\_biogen\_oa\_cm} configuration --- the model selected for cross-baseline comparison in Section~\ref{sec:comp_with_baselines} --- leads the win-count and best-readout columns on the ExpansionRX source, and is tied for the best readout-averaged classification score on the Biogen source.


\FloatBarrier
\newpage
\makeatletter
\if@preprint\else
  \section*{NeurIPS Paper Checklist}

\begin{enumerate}

\item {\bf Claims}
    \item[] Question: Do the main claims made in the abstract and introduction accurately reflect the paper's contributions and scope?
    \item[] Answer: \answerYes{} 
    \item[] Justification: The abstract and introduction accurately summarize the paper's main contributions: the Contrastive KERMT pretraining framework, the probabilistic integration of cMIM with chemistry-specific self-supervised tasks, task-specific fine-tuning heads, and evaluation on Biogen, ExpansionRX, and ChEMBL-MT ADME benchmarks. The claims are supported by the method description, baseline comparisons, and ablation studies, and are appropriately scoped to ADME-focused molecular property prediction rather than broader molecular modeling tasks.

\item {\bf Limitations}
    \item[] Question: Does the paper discuss the limitations of the work performed by the authors?
    \item[] Answer: \answerYes{} 
    \item[] Justification: A Limitations paragraph (Section~\ref{sec:limitations}) discusses the ADME-only evaluation scope, use of a single KERMT graph-transformer backbone, label-free/transductive corpus-adaptation setting for some pretraining configurations, and compute burden from large-scale pretraining.

\item {\bf Theory assumptions and proofs}
    \item[] Question: For each theoretical result, does the paper provide the full set of assumptions and a complete (and correct) proof?
    \item[] Answer: \answerNA{} 
    \item[] Justification: The paper introduces and evaluates a pretraining objective but does not present new theoretical theorems or proofs. The probabilistic objective definitions are provided in Section~\ref{sec:mim-loss}.

    \item {\bf Experimental result reproducibility}
    \item[] Question: Does the paper fully disclose all the information needed to reproduce the main experimental results of the paper to the extent that it affects the main claims and/or conclusions of the paper (regardless of whether the code and data are provided or not)?
    \item[] Answer: \answerYes{} 
    \item[] Justification: We provide all the code used to run the experiments reported in this manuscript as a zip file at the time of manuscript submission. We will release all the code and pretrained \modelname checkpoints in a public repository at publication time. We provide details of all model hyperparameters used to run the experiments and detail the splits used for datasets. We have included all required information to reproduce the results to the best of our knowledge.

\item {\bf Open access to data and code}
    \item[] Question: Does the paper provide open access to the data and code, with sufficient instructions to faithfully reproduce the main experimental results, as described in supplemental material? 
    \item[] Answer: \answerYes{} 
    \item[] Justification: The paper will release the code used to run Contrastive KERMT model and reproduce all experiments at publication time. Well-documented code in a zip file will be included with the submission. We will also release all the data used for pretraining, MolMIM-generated molecules, and data splits used for fine-tuning at publication time.

\item {\bf Experimental setting/details}
    \item[] Question: Does the paper specify all the training and test details (e.g., data splits, hyperparameters, how they were chosen, type of optimizer) necessary to understand the results?
    \item[] Answer: \answerYes{} 
    \item[] Justification: Yes, the paper provides train/test data splits in Table~\ref{tab:dataset-overview}. Hyperparameters for the Contrastive KERMT model are detailed in Section~\ref{sec:model-variants}, and implementation details including training budgets, fine-tuning protocols, baselines, and statistical testing are provided in Appendix~\ref{app:implementation-details}.

\item {\bf Experiment statistical significance}
    \item[] Question: Does the paper report error bars suitably and correctly defined or other appropriate information about the statistical significance of the experiments?
    \item[] Answer: \answerYes{} 
    \item[] Justification: This paper reports error bars  or confidence intervals for all experiments. They are available in Tukey HSD plots and bar plots (Figures \ref{fig:comp_with_baselines-biogen}, \ref{fig:comp_with_baselines-exprx}, \ref{fig:comp_with_baselines-chembl-mt}, \ref{fig:finetune-biogen}, \ref{fig:finetune-openadmet}, \ref{fig:finetune-chemblmt}). In multiple comparisons, ANOVA followed by pairwise Tukey's Honestly Significant Difference was used. When comparing mean of two methods, statistical significance was determined based on student's t-test for means. In bar plots, experiments statistically significant differences are shown using '*'.

\item {\bf Experiments compute resources}
    \item[] Question: For each experiment, does the paper provide sufficient information on the computer resources (type of compute workers, memory, time of execution) needed to reproduce the experiments?
    \item[] Answer: \answerYes{} 
    \item[] Justification: Hardware (8 NVIDIA A100 GPUs on an internal compute cluster, NCCL DDP), per-GPU and effective global batch sizes, and approximate per-run wall-clock cost for each pretraining configuration as well as fine-tuning are reported in Appendix~\ref{app:implementation-details} (Hardware and compute paragraph).
    
\item {\bf Code of ethics}
    \item[] Question: Does the research conducted in the paper conform, in every respect, with the NeurIPS Code of Ethics \url{https://neurips.cc/public/EthicsGuidelines}?
    \item[] Answer: \answerYes{} 
    \item[] Justification: The paper’s core claims are reasonably supported by the manuscript. The abstract states the method, benchmarks, and reported gains, and the experiments compare Contrastive KERMT across Biogen, ExpansionRX, and ChEMBL-MT with reported improvements and ablations. The work does not appear to involve human subjects, crowdsourcing, or personally identifiable information. The paper provides a meaningful amount of experimental detail: pretraining schedules, cMIM temperature, fine-tuning architecture, number of seeds/runs, baselines, and statistical testing are described in Appendix \ref{app:implementation-details}.

\item {\bf Broader impacts}
    \item[] Question: Does the paper discuss both potential positive societal impacts and negative societal impacts of the work performed?
    \item[] Answer: \answerYes{} 
    \item[] Justification: This work targets ADME property prediction, an upstream component of drug-discovery pipelines; improved in-silico ADME models can accelerate candidate prioritization and reduce reliance on animal-using assays. Model predictions are not a substitute for experimental validation, and naïve reliance on them without wet-lab confirmation could misallocate medicinal-chemistry effort, though this risk is well-known in the QSAR/ADME community and is not specifically increased by our contribution.
    
\item {\bf Safeguards}
    \item[] Question: Does the paper describe safeguards that have been put in place for responsible release of data or models that have a high risk for misuse (e.g., pre-trained language models, image generators, or scraped datasets)?
    \item[] Answer: \answerNA{}
    \item[] Justification: The paper does not pose high risks for misuse and does not release any new data or high-risk models. It only uses publicly available datasets that have already been released, so additional safeguards for responsible release are not applicable.

\item {\bf Licenses for existing assets}
    \item[] Question: Are the creators or original owners of assets (e.g., code, data, models), used in the paper, properly credited and are the license and terms of use explicitly mentioned and properly respected? 
    \item[] Answer: \answerYes{} 
    \item[] Justification: Licenses for fine-tuning datasets can be found in Table \ref{tab:dataset-overview}. Licenses for pretraining datasets are indicated in Appendix \ref{app:datasets}. License information for the KERMT and baseline implementations is provided in Appendix~\ref{app:implementation-details}.

\item {\bf New assets}
    \item[] Question: Are new assets introduced in the paper well documented and is the documentation provided alongside the assets?
    \item[] Answer: \answerYes{} 
    \item[] Justification: Well-documented code with model implementation is included as a zip file with the submission for reviewers. The code, pretrained \modelname checkpoints, data splits, and MolMIM-generated molecules will be released to the public at the time of publication. 

\item {\bf Crowdsourcing and research with human subjects}
    \item[] Question: For crowdsourcing experiments and research with human subjects, does the paper include the full text of instructions given to participants and screenshots, if applicable, as well as details about compensation (if any)? 
    \item[] Answer:\answerNA{} 
    \item[] Justification: This paper does not include any crowdsourcing experiments or research with human subjects.

\item {\bf Institutional review board (IRB) approvals or equivalent for research with human subjects}
    \item[] Question: Does the paper describe potential risks incurred by study participants, whether such risks were disclosed to the subjects, and whether Institutional Review Board (IRB) approvals (or an equivalent approval/review based on the requirements of your country or institution) were obtained?
    \item[] Answer: \answerNA{}
    \item[] Justification: The paper does not involve crowdsourcing, user studies, interviews, surveys, experiments with human participants, or collection of human-subject data. Therefore, IRB approval or equivalent review is not applicable.

\item {\bf Declaration of LLM usage}
    \item[] Question: Does the paper describe the usage of LLMs if it is an important, original, or non-standard component of the core methods in this research? Note that if the LLM is used only for writing, editing, or formatting purposes and does \emph{not} impact the core methodology, scientific rigor, or originality of the research, declaration is not required.
    \item[] Answer: \answerNA{} 
    \item[] Justification:  LLMs were used only for writing, editing, and formatting the manuscript. They were not used as an important, original, or non-standard component of the core research methodology, experiments, analysis, or scientific contributions. Therefore, a declaration of LLM usage is not applicable.

\end{enumerate}

\fi
\makeatother

\end{document}